\documentclass{article}

\usepackage{iclr2027_conference}
\iclrfinalcopy
\usepackage{iftex}
\ifPDFTeX
  \usepackage{times}
\else
  \usepackage{fontspec}
\fi

\usepackage{amsmath,amssymb,amsthm,mathtools}
\usepackage{array}
\usepackage{booktabs}
\usepackage{float}
\usepackage{graphicx}
\usepackage{microtype}
\usepackage{multirow}
\usepackage{placeins}
\usepackage{url}
\usepackage{xcolor}
\usepackage{colortbl}
\usepackage[hidelinks]{hyperref}


\newtheorem{proposition}{Proposition}

\theoremstyle{remark}

\title{TV-Regulated OPD: Direction Matters in On-Policy Distillation}
\author{%
  Han Xiao\textsuperscript{1 *}\quad
  Yifan Niu\textsuperscript{1 *}\quad
  Dongyi Liu\textsuperscript{1}\quad
  Chang Luo\textsuperscript{2}\quad
  Jia Li\textsuperscript{1,3 $\dagger$}\\[0.45em]
  \normalfont\normalsize
  \textsuperscript{1}The Hong Kong University of Science and Technology (Guangzhou)\\
  \textsuperscript{2}The University of Edinburgh\\
  \textsuperscript{3}The Hong Kong University of Science and Technology
}

\makeatletter
\def\@maketitle{%
  \vbox{\hsize\textwidth
    {\LARGE\scshape \@title\par}
    \vskip 0.3in minus 0.1in
    {\centering\large\bfseries \@author\par}
    \vskip 0.3in minus 0.1in
  }%
}
\makeatother

\hypersetup{
  pdftitle={TV-Regulated OPD: Direction Matters in On-Policy Distillation},
  pdfauthor={Han Xiao, Yifan Niu, Dongyi Liu, Chang Luo, Jia Li}
}

\begin{document}

\maketitle
\lhead{Preprint}
\pagestyle{fancy}
\begingroup
\renewcommand{\thefootnote}{\fnsymbol{footnote}}
\footnotetext[1]{Equal contribution.}
\footnotetext[2]{Correspondence to Jia Li (jialee@ust.hk).}
\endgroup

\begin{abstract}
On-Policy Distillation (OPD) facilitates the transfer of knowledge from domain expert to student in the post-training phase of Large Language Models (LLMs). However, the supervision signals in mainstream OPD methods suffer from high variance and noise which is generally instable during training. 
In this work, we systematically investigated what really matters to the performance and the fundamental mechanisms behind the instability during training.
We found that retaining only the sign of token-level advantages is sufficient to achieve the performance comparable to standard OPD. Meanwhile, smoother and bounded advantages can stabilize the training process without sacrificing its performance.
These motivated us to shape the advantages using the Total Variation (TV) and propose a robust TV regulated On-Policy Distillation (TV-OPD) method.
Benefiting from the bounded and diminished advantages, TV-OPD exhibits stable training dynamics and steady late-stage performance.
We conducted comprehensive experiments and found that, across various settings, TV-OPD consistently achieved better performance and lower variance in the late-stage of training. The code is available at \href{https://github.com/O617/OPD}{https://github.com/O617/OPD}.

\end{abstract}

\section{Introduction}
\label{sec:introduction}

On-Policy Distillation (OPD) has emerged as a key method for the efficient transfer of knowledge from teacher to student in the post-training of large models~\citep{lu2025opd,li2026rethinkingopd,song2026surveyonpolicydistillationlarge}. Unlike offline distillation~\citep{agarwal2023gkd}, OPD provides dense, fine-grained supervision signals directly on trajectories sampled from the student policy, making it particularly effective for fine-grained policy optimization during post-training. Leading industries such as Qwen3~\citep{yang2025qwen3}, GLM~\citep{glm5team2026glm5}, and Mimo~\citep{xiao2026mimov2flash} have integrated OPD into their training pipelines and reported its significant impact.

Mainstream OPD methods employ Reverse Kullback-Leibler (KL) divergence as the optimization objective, efficiently propagating supervision signals via token-level advantages~\citep{lu2025opd,li2026rethinkingopd}. This advantage can be decomposed into sign component and magnitude component where sign determines whether a token should be encouraged or suppressed and magnitude influences the size of the corresponding gradient. However, the absolute value of the advantage is an unbounded and highly non-linear logarithmic function. Existing studies have also observed that supervision signals exhibit extremely high variance, making OPD training unstable~\citep{oh2026vopd,zhao2026poweropd,wang2026demystifying}.

To stabilize OPD training, existing approaches smooth and bound the advantage. Common techniques include truncating extreme values and remapping values to a bounded range~\citep{wang2026demystifying,zhao2026poweropd,yu2026tide}. These methods share a common premise that token-level magnitude information is crucial for the teacher to effectively supervise the student. Therefore, any transformation aimed at stabilizing training must preserve this information. However, we found that this hypothesis does not necessarily hold. To investigate the role of magnitude information, we retained the sign of the advantage while removing the magnitude information to varying degrees. We observed that the model continued to achieve stable improvements even as token-level magnitude information was progressively erased or even shuffled, indicating that magnitude information has a relatively minor impact on OPD training.

\begin{figure}[!t]
  \centering
  \setlength{\abovecaptionskip}{0pt}
  \setlength{\belowcaptionskip}{0pt}
  \includegraphics[width=\textwidth,trim=7bp 70bp 9bp 48bp,clip]{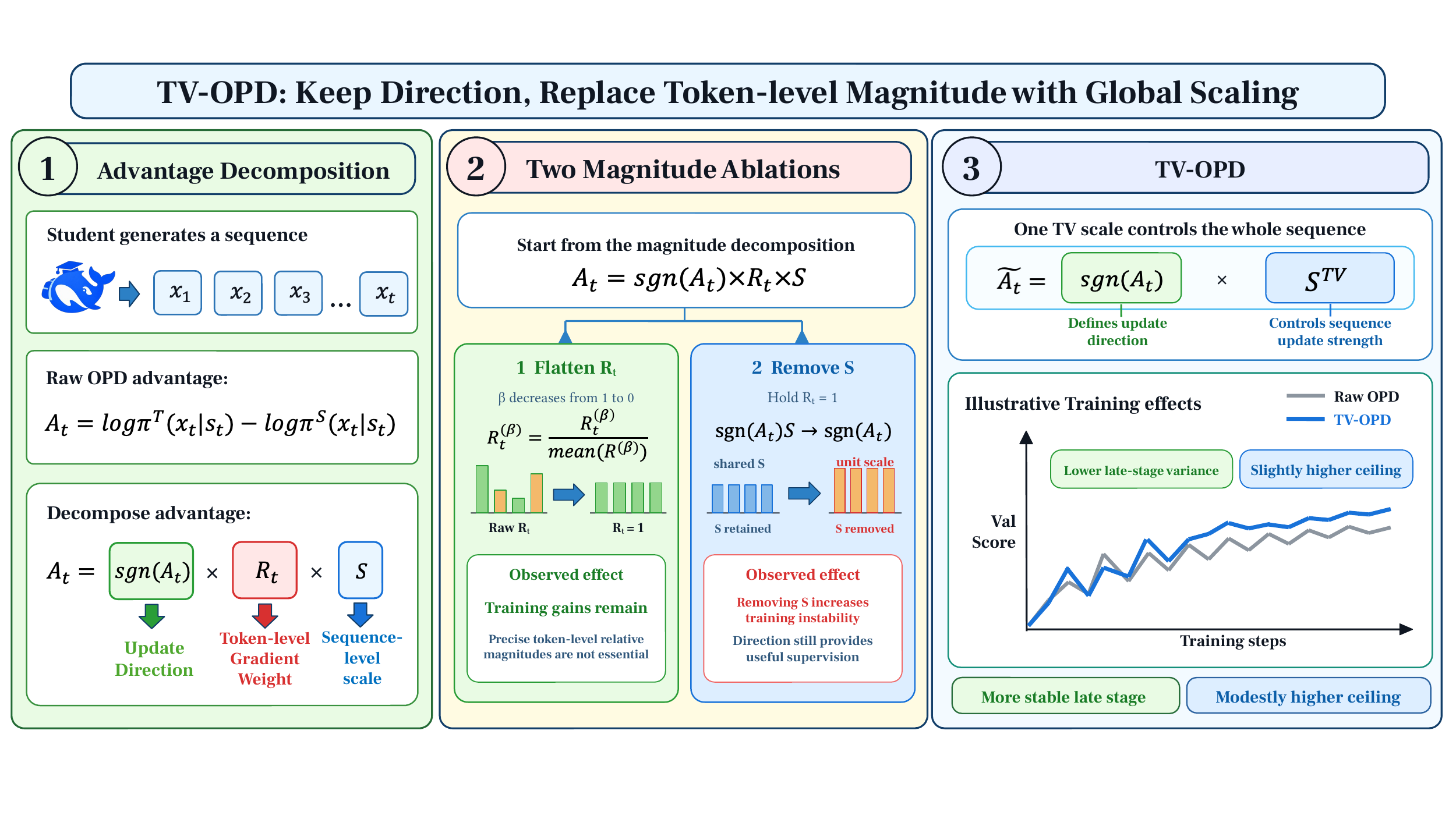}
  \caption{Overview of TV-OPD: magnitude ablations motivate retaining token-level update directions and using one TV-based sequence scale for stable late-stage training.}
  \label{fig:tvopd-overview}
  \vspace{-12pt}
\end{figure}

Inspired by the aforementioned observations, we propose a more robust TV-regulated OPD (TV-OPD) method. Since magnitudes with extremely high variance can compromise training stability while the magnitude information itself has a relatively minor impact on the training process, we eliminated token-level magnitudes. Furthermore, to facilitate better convergence, we employ a scaling factor that decreases as the student model approaches the teacher model, thereby controlling the update strength at each step. Figure~\ref{fig:tvopd-overview} summarizes the advantage decomposition, the two motivating magnitude ablations, and the resulting TV-OPD update. We conducted experiments across multiple teacher-pair and benchmark settings. Experimental results demonstrate that TV-OPD achieves a higher training performance ceiling in both 1.5B and 8B settings, while also ensuring greater stability during training.

\section{Related Work}
\label{sec:related-work}

On-Policy Distillation (OPD) optimizes student model by providing dense teacher supervision on trajectories generated by the student model itself to have better distribution alignment and performance~\citep{agarwal2023gkd,lu2025opd}. In early works, MiniLLM pioneered the on-policy distillation by using a reverse KL divergence objective enforce mode-seeking behaviors~\citep{gu2024minillm}. GKD established a unified distillation framework supporting various sampling strategies and divergence metrics~\citep{agarwal2023gkd}. Recent research classifies OPD supervision signals into self-play, outcome-based and logit-based rewards~\citep{song2026surveyonpolicydistillationlarge}. Self-play methods allow the model to act as both teacher and student and outcome-based rewards utilize sequence-level reward signals. Our research focus on Logit-based methods which provide dense supervision by comparing the token-level probability distributions of the teacher and the student. These methods use the log probabilities difference between the teacher and the student on sampled tokens as token-level advantage which is high-variance and has extreme values~\citep{li2026rethinkingopd,oh2026vopd,zhao2026poweropd}. The challenge of effectively and stably leveraging the information containing in it has become the focus in subsequent OPD research.

To enhance the stability and effectiveness of OPD training, existing methods transform the token-level advantage to reduce its extreme values and variance. PowerOPD computes the difference between the power-transformed token probabilities of the teacher and the student to construct a bounded supervision signal with the consistent direction~\citep{zhao2026poweropd}. ClipOPD directly truncates the token-level advantage within a fixed range, limiting the impact of extreme values on gradient updates~\citep{wang2026demystifying}. TIDE applies a bounded Hellinger transformation to "student-excess" tokens, mapping the originally unbounded negative advantage to a bounded signal—while supplementing supervision for "student-deficit" tokens via teacher top-$K$ injection~\citep{yu2026tide}. vOPD subtracts a control-variate baseline derived from token-level reverse KL from the raw advantage, reducing estimation variance while preserving the expected gradient~\citep{oh2026vopd}. Despite their differing specific forms, these methods generally operate on the assumption that token-level magnitude information is crucial for the teacher to effectively supervise the student, and they transform this information to stabilize training. Our observations, however, indicate that such token-level magnitude information is not necessarily essential for OPD training.

\section{Preliminary}
\label{sec:preliminary}
\label{sec:direction-magnitude}
\textbf{Notation.} Let $x = \{x_1, \ldots, x_n\}$ denote the input query and $y = \{y_1, \ldots, y_m\}$ denote the response. The token prefix prior to step $t$ is defined as $y_{<t} = (y_1, \ldots, y_{t-1})$. We consider a student model and a teacher model that share a vocabulary $\mathcal{V}$, denoting their respective predictive distributions for the next token which conditioned on $x$ and $y_{<t}$ as $\pi_\theta$ and $\pi_T$. We use $y \sim \pi_\theta(\cdot \mid x)$ to represent a response sampled autoregressively from the student model, and $\mathcal{D}_x =\{x^{(i)}\}_{i=1}^N$ to represent the corresponding dataset of queries.

\subsection{On-Policy Distillation}
On-Policy Distillation (OPD) computes supervision signals based on trajectories sampled from the current student model $\pi_\theta$. Given a prompt $x \sim \mathcal{D}_x$, the student model generates a response $\hat{y}=(\hat{y}_1,\ldots,\hat{y}_T)\sim\pi_\theta(\cdot\mid x)$. Subsequently, we concatenate the prompt and response into a complete sequence and feed it into both the student and teacher models for a forward pass. This yields the predictive distributions for the next token at each step—$\pi_\theta(v\mid x,\hat{y}_{<t})$ and $\pi_T(v\mid x,\hat{y}_{<t})$, where $v\in\mathcal{V}$. In this work, we base our study on the sample-based OPD method proposed by Thinking Machines Lab~\citep{lu2025opd}. This method employs reverse KL divergence as the distillation objective and uses Monte Carlo estimation based on tokens actually sampled by the student model; it represents a lightweight implementation commonly used in existing OPD research~\citep{agarwal2023gkd,li2026rethinkingopd}. During a policy update, let $\pi_{\theta_{\mathrm{old}}}$ denote the student policy that generates the trajectory; the advantage corresponding to the $t$-th token is defined as
\begin{equation}
\hat{A}_t
=
\log \pi_T(\hat{y}_t\mid x,\hat{y}_{<t})
-
\log \pi_{\theta_{\mathrm{old}}}(\hat{y}_t\mid x,\hat{y}_{<t}).
\label{eq:raw_advantage}
\end{equation}
Since $\hat{y}_t\sim\pi_{\theta_{\mathrm{old}}}(\cdot\mid x,\hat{y}_{<t})$,
its conditional expectation satisfies
\begin{equation}
\mathbb{E}_{\hat{y}_t\sim\pi_{\theta_{\mathrm{old}}}}
[\hat{A}_t]
=
-
D_{\mathrm{KL}}
\!\left(
\pi_{\theta_{\mathrm{old}}}(\cdot\mid x,\hat{y}_{<t})
\,\|\,
\pi_T(\cdot\mid x,\hat{y}_{<t})
\right),
\end{equation}
thus $\hat{A}_t$ can be viewed as a single-sample estimate of the negative reverse KL divergence.

During the optimization phase, we employ importance sampling to utilize the supervisory signal obtained under the generation policy $\pi_{\theta_{\mathrm{old}}}$ to update the current policy $\pi_\theta$. Defining the probability ratio as
$r_t(\theta)
=
\frac{
\pi_\theta(\hat{y}_t\mid x,\hat{y}_{<t})
}{
\pi_{\theta_{\mathrm{old}}}(\hat{y}_t\mid x,\hat{y}_{<t})
},$
the OPD surrogate objective employing PPO-style clipping
\citep{schulman2017ppo} is given by
\begin{equation}
\mathcal{J}_{\mathrm{OPD}}(\theta)
=
\mathbb{E}_{x\sim\mathcal{D}_x,\,
\hat{y}\sim\pi_{\theta_{\mathrm{old}}}}
\min\left(
r_t(\theta)\hat{A}_t,\,
\operatorname{clip}
\big(r_t(\theta),1-\epsilon,1+\epsilon\big)\hat{A}_t
\right).
\label{eq:opd_objective}
\end{equation}
Optimization process maximized $\mathcal{J}_{\mathrm{OPD}}$ to gradually align the student's policy with the teacher's policy on the states visited by the student.

\subsection{Advantage Decomposition}

According to Equation \eqref{eq:opd_objective}, the token-level advantage $\hat{A}_t$ directly determines the contribution of each sampled token to the policy update. It can be naturally decomposed into sign and magnitude components:
\begin{equation}
\hat{A}_t
=
\operatorname{sgn}(\hat{A}_t)
\cdot |\hat{A}_t|.
\label{eq:adv_decomposition}
\end{equation}

The sign of $\hat{A}_t$ determines whether the sampled token $\hat{y}_t$ is encouraged or suppressed, thereby controlling the direction of the token's update; meanwhile, its magnitude determines the extent to which the token is encouraged or suppressed, thereby controlling the intensity of the update. The student model is encouraged to increase the probability of generating $\hat{y}_t$ when $\hat{A}_t > 0$ and to decrease this probability when $\hat{A}_t < 0$. The magnitude $|\hat{A}_t|$ effectively assigns a weight to each token, modulating the strength of the gradient associated with that token.

\section{Observation: Direction and Magnitude in OPD}
\label{sec:observation}
\label{sec:magnitude-ablations}

Equation~\eqref{eq:adv_decomposition} decomposes the advantage into a sign and
a magnitude. However, the magnitude itself still plays two distinct roles:
(1) The magnitude associated with each token determines the weight of its
gradient within the sequence-level gradient, thereby influencing the
gradient's composition. (2) The token-level magnitudes determine the sequence-level magnitude, thereby controlling the strength of a single update. To investigate the role of each component, we further decompose the advantage. Taking a sampled sequence of length $T$, we express the advantage
of each token as
\begin{equation}
\hat{A}_t
= \operatorname{sgn}(\hat{A}_t)\,R_tS.
\label{eq:fine_grained_adv_decomposition}
\end{equation}
$R_t = \frac{|\hat{A}_t|}{S}$ represents the relative magnitude of token $t$, determining the
proportion of that token's gradient within the sequence-level gradient, and
$S = T^{-1}\sum_{j=1}^{T}|\hat{A}_j|$ is a sequence-level scale that influences the strength of the update. Then we train the model after removing each of these two components from the advantage and observe the resulting training performance.

\subsection{Observation 1: Precise Token-Level Relative Magnitudes Are Not Essential}

To gradually remove the information conveyed by $R_t$, we define a transformation as follows:
\begin{equation}
R_t^{(\beta)}
= \frac{R_t^\beta}{T^{-1}\sum_{j=1}^{T}R_j^\beta},
\qquad
\beta \in [0, 1]
\label{eq:power_transformation}
\end{equation}
and apply it to the original advantage as
$
\tilde A_t^{(\beta)}
= \operatorname{sgn}(\hat A_t)R_t^{(\beta)}S.
\label{eq:power_transformed_advantage}
$
We gradually decrease $\beta$ from 1 to 0, which progressively flattens $R_t$ toward 1 while preserving its sign and $S$. We selected $\beta$ values of 0.4 and 0.8 to observe the results, ultimately obtaining four experimental groups: Raw, Power ($\beta=0.4$), Power ($\beta=0.8$), and Sequence-Constant. Additionally, we shuffle the $R_t$ values of individual tokens within each sequence which completely disrupts the token-level relative magnitude in the sequence while retaining the original signs and sequence-level scaling characteristics.

\begingroup
\setlength{\intextsep}{4pt}
\setlength{\abovecaptionskip}{4pt}
\setlength{\belowcaptionskip}{0pt}
\begin{figure}[H]
  \centering
  \includegraphics[width=0.99\textwidth]{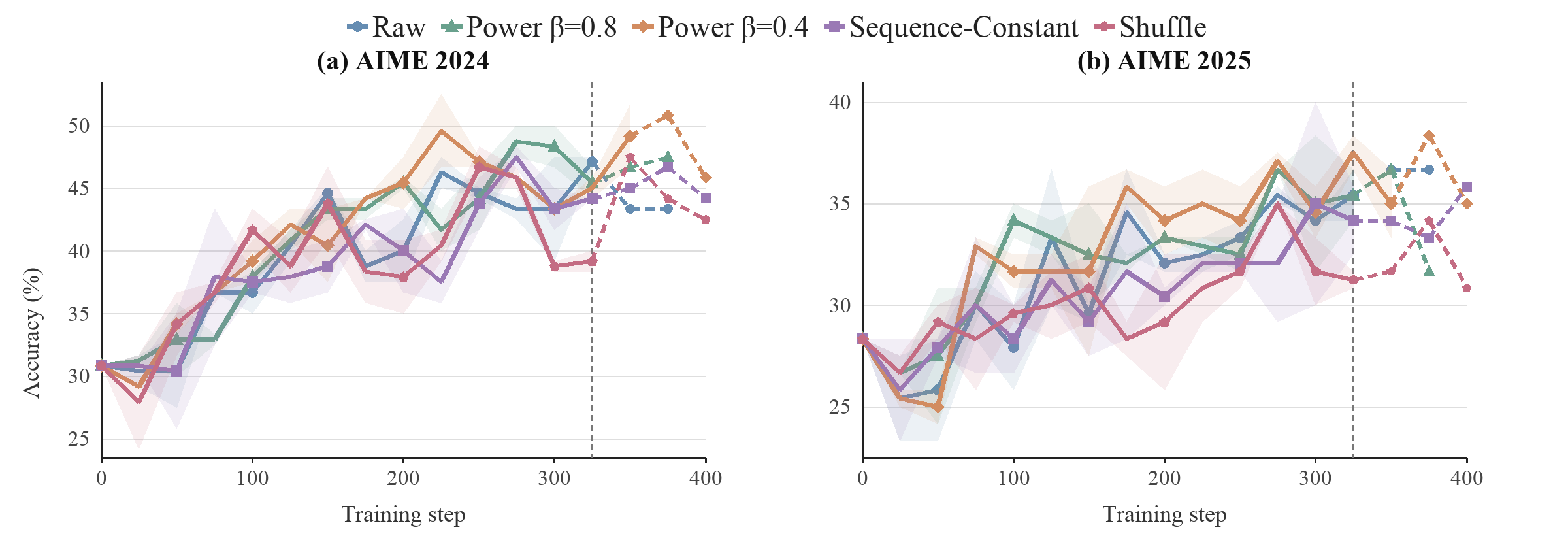}
  \caption{Relative-magnitude ablations on the JustRL-DeepSeek-1.5B pair
  (mean $\pm$ one standard deviation across runs; dashed tails denote
  reduced coverage).}
  \label{fig:relative-magnitude-ablation}
\end{figure}
\endgroup

Figure~\ref{fig:relative-magnitude-ablation} shows that weakening
relative magnitudes does not degrade training. Power-$0.4$ and Power-$0.8$
obtain two-benchmark trajectory averages of $36.71\%$ and $36.34\%$, versus
$35.40\%$ for Raw. Shuffle remains competitive at $34.35\%$. Thus, precise
token-wise magnitude may not be essential.

\subsection{Observation 2: Direction Alone Retains Effective Supervision}

To assess the impact of $S$, we remove $S$ from the Sequence-Constant group to
obtain the Sign group:
$
\tilde{A}^{\mathrm{Sign}}_t
= \operatorname{sgn}(\hat A_t).
\label{eq:global_scale_ablation}
$
This comparison holds the token-level allocation fixed at $R_t=1$ and changes
only whether the sequence-level scale is retained.

\begingroup
\setlength{\intextsep}{4pt}
\setlength{\abovecaptionskip}{4pt}
\setlength{\belowcaptionskip}{0pt}
\begin{figure}[H]
  \centering
  \includegraphics[width=0.99\textwidth]{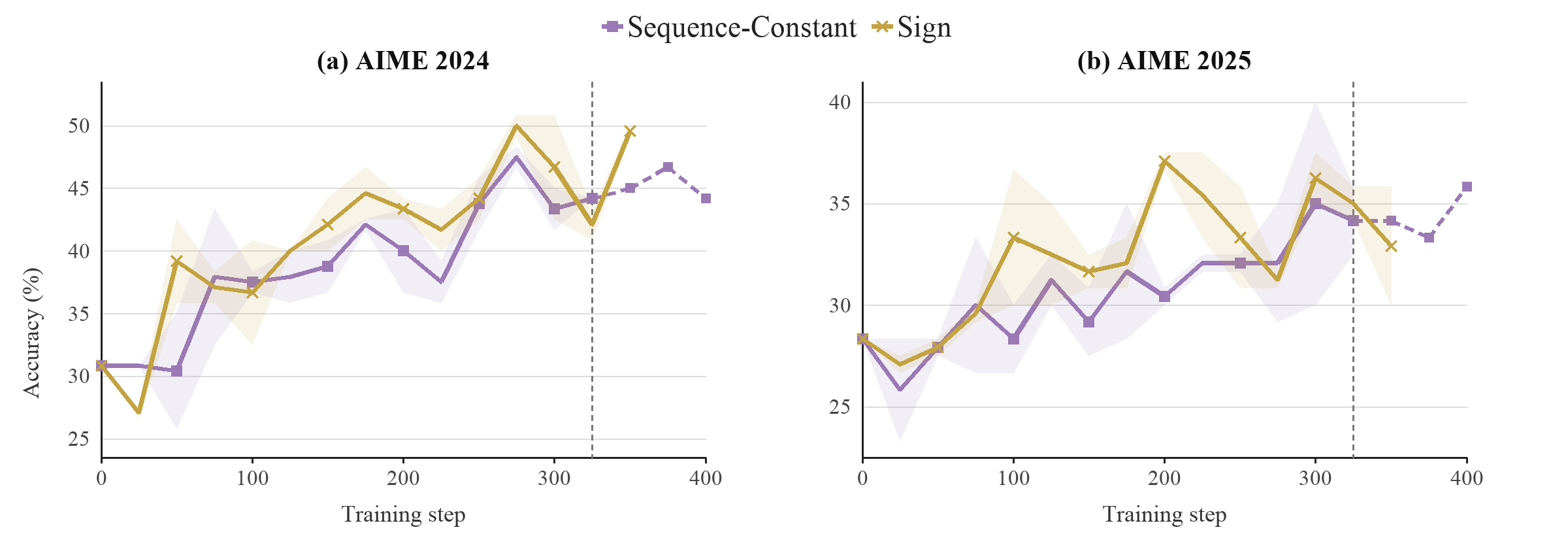}
  \caption{Sequence-scale ablation on the same pair. Sequence-Constant retains
  $S$, whereas Sign uses direction alone; plotting conventions follow
  Figure~\ref{fig:relative-magnitude-ablation}.}
  \label{fig:sequence-scale-ablation}
\end{figure}
\endgroup

Figure~\ref{fig:sequence-scale-ablation} shows that Sign raises the common-horizon
two-benchmark average from $34.67\%$ to $36.29\%$. It also improves
best-over-training accuracy from $47.50\pm1.18$ to $50.00\pm1.18$ on AIME 2024
and from $36.25\pm5.30$ to $37.50\pm0.00$ on AIME 2025
(Appendix~\ref{app:magnitude-ablations}).

Direction alone therefore provides useful supervision, but removing $S$ also
discards discrepancy-dependent attenuation. This motivates TV-OPD: token signs
set local update directions, while aggregate TV controls their shared strength.
Section~\ref{sec:experiments} evaluates this design on a second model pair.

\section{Method: TV-Regulated OPD}
\label{sec:method}
\label{sec:tv-theory}
\label{sec:scheduled-tv}

In this section, we introduce the Total Variation-regularized OPD (TV-OPD) method. Based on our observations, reducing the variance of token-level magnitudes and gradually decreasing the update step size as training converges helps enhance the training stability of OPD. We first remove the relative magnitude component. Then we use the Total Variation (TV) distance between the teacher and student models to measure training progress and dynamically control the global update strength.

\subsection{Direction-Preserving Advantage}

Our Observation 1 indicates that removing the precise token-level relative magnitude—which is precisely the source of the high variance in token-level magnitudes—has no impact on effective distillation. This motivated us to remove $R_t$ from Equation \eqref{eq:fine_grained_adv_decomposition}, retaining only the sign and the sequence-level magnitude $S$:
\begin{equation}
\tilde A_t=S\cdot\operatorname{sgn}(\hat A_t),
\label{eq:direction-preserving-advantage}
\end{equation}
where $\operatorname{sgn}(0)=0$. Considering a state $s=(x,\hat y_{<t})$ visited by the student model, let $p=\pi_T(\cdot\mid s)$ and $q=\pi_{\theta_{\mathrm{old}}}(\cdot\mid s)$ denote the distributions of the teacher model and the sampling student model, respectively. Their conditional total variation distance is given by
\begin{equation}
D_{\mathrm{TV}}(p,q)
=\frac12\sum_{v\in\mathcal V}|p(v)-q(v)|.
\label{eq:tv-definition}
\end{equation}
Since the logarithm function is strictly monotonically increasing, we have
$\operatorname{sgn}(\log p(a)-\log q_\theta(a))
=\operatorname{sgn}(p(a)-q_\theta(a))$, where
$q_\theta=\pi_\theta(\cdot\mid s)$ represents the current student model. Differentiating the TV distance and applying the score identity yields:
\begin{equation}
\begin{aligned}
-2\nabla_\theta D_{\mathrm{TV}}(p,q_\theta)
&=\sum_{a\in\mathcal V}\operatorname{sgn}(p(a)-q_\theta(a))
\nabla_\theta q_\theta(a)\\
&=
\mathbb E_{a\sim q_\theta}\!\left[
\operatorname{sgn}\!\left(\log\frac{p(a)}{q_\theta(a)}\right)
\nabla_\theta\log q_\theta(a)\right].
\end{aligned}
\label{eq:tv-gradient}
\end{equation}
Thus, training using this transformed advantage is equivalent to optimizing the TV distance between the student and the teacher based on trajectories sampled by the student. Furthermore, this equality can be interpreted from the perspective of subgradients (see Appendix~\ref{app:theory-proofs}). This additional connection motivates us to use the TV distance to regulate the shared scale factor.

\subsection{TV-Based Sequence Scaling}

To use TV for global regulation, we estimate its value from the sampled-token log-probabilities already available in OPD.
Because the teacher $p$ and rollout student $q$ are normalized, the student's
total excess probability equals its total deficit. Hence,
\begin{equation}
  D_{\mathrm{TV}}(p,q)
  =\sum_{v\in\mathcal V}[q(v)-p(v)]_+
  =\mathbb E_{a\sim q}\!\left[1-\frac{p(a)}{q(a)}\right]_+,
  \label{eq:tv-one-sided-identity}
\end{equation}
where $[z]_+=\max(z,0)$. Since
$\hat A_t=\log p(\hat y_t)-\log q(\hat y_t)$, an online estimate is
\begin{equation}
  \widehat d_t=[1-\exp(\hat A_t)]_+,
  \qquad
  \mathbb E[\widehat d_t\mid s]=D_{\mathrm{TV}}(p,q).
  \label{eq:tv-estimator}
\end{equation}
It reuses the original sampled-token log-probabilities, requires no additional
teacher scoring, and lies in $[0,1]$ with conditional variance at most $1/4$.
At training step $k$, we pool these estimates across active tokens,
microbatches, and workers to obtain $\widehat D_k$. This estimates average
conditional TV along student rollouts. Fixed-horizon averaging is unbiased;
variable-length pooling is a consistent ratio estimator under independent
rollouts and masks determined before the current token is sampled, but need
not be unbiased at finite batch size. Appendix~\ref{app:tv-estimation}
provides the statistical details.

The sequence-level $S$ retains raw log-ratio magnitude only as an aggregate.
TV-OPD replaces this scale with a shared, TV-guided coefficient $c_k$, so
that global supervision strength responds to teacher--student discrepancy
without restoring relative token magnitudes. We smooth the online estimates
using an exponential moving average,
$\bar D_k=\beta\bar D_{k-1}+(1-\beta)\widehat D_k$, and define
\begin{equation}
  A_{t,k}^{\mathrm{TV}}=c_k\operatorname{sgn}(\hat A_{t,k}),
  \qquad
  c_k=\operatorname{clip}\!\left[
    \left(\frac{\bar D_{k-1}+\epsilon}{D_{\mathrm{ref}}+\epsilon}\right)^{\!\alpha},
    c_{\min},1\right].
  \label{eq:scheduled-tv-advantage}
\end{equation}
Here $D_{\mathrm{ref}}$ is fixed to the first valid step's estimate, which also
initializes the moving average. Initially $c_k=1$; subsequent coefficients
use only the previous step's moving average. The constant $\epsilon>0$
stabilizes the ratio, $\alpha>0$ controls attenuation, and
$0<c_{\min}\le1$ sets a floor. As smoothed TV falls below its reference,
$c_k$ decreases; larger $\alpha$ produces stronger attenuation.

We substitute $A_{t,k}^{\mathrm{TV}}$ into Eq.~\eqref{eq:opd_objective},
keeping $c_k$ detached and fixed across the step. Its positive, shared value
regulates the pre-optimizer gradient strength while preserving all local
signs. Using a historical estimate avoids coupling the coefficient to the
current sampled gradient and preserves the expected TV direction whenever
the base sign signal is unbiased under the chosen token weighting
(Appendix~\ref{app:tv-scheduler}). TV-OPD thus uses direction for token-level
supervision and online discrepancy for global scaling.

\section{Experiments}
\label{sec:experiments}

\subsection{Setup}

\paragraph{Model pairs.}
We evaluate two same-family teacher--student pairs. In the 8B setting, the
teacher is Qwen3-8B and the student is Qwen3-8B-Base after supervised
fine-tuning on 400K OpenThoughts examples; OPD uses DeepMath-103K prompts
\citep{yang2025qwen3,guha2025openthoughts,he2025deepmath}. In the 1.5B
setting, the teacher is JustRL-DeepSeek-1.5B, the student starts from
DeepSeek-R1-Distill-Qwen-1.5B, and OPD uses DAPO-Math-17K
\citep{deepseekai2025r1,he2025justrl,yu2025dapo}. In both settings, teacher
and student use the same tokenizer. This removes cross-tokenizer alignment as
a confounding factor.

\paragraph{Baselines.}
We compare against the untrained student, Raw OPD~\citep{lu2025opd},
ClipOPD~\citep{wang2026demystifying}, PowerOPD~\citep{zhao2026poweropd},
and vOPD~\citep{oh2026vopd}. These baselines cover
the untransformed sampled-token objective, direct clipping, bounded power
shaping, and control-variate variance reduction, respectively. All methods use the same model pair, OPD prompts, rollout
budget, optimizer, and evaluation protocol; only the method-specific
distillation signal changes. Appendix~\ref{app:baselines} gives the detailed
definitions.

\paragraph{Training and evaluation.}
Each update uses 64 prompts and one on-policy rollout per prompt. We train with
AdamW~\citep{loshchilov2019adamw} at learning rate $10^{-6}$. We evaluate
periodically on AIME 2024,
AIME 2025, AIME 2026, AMC 2023, HMMT 2025, and MATH-500 with four responses
per problem for training-dynamics analyses. For the main comparison, we use
16 responses per problem and report mean@16~\citep{maa2026amc,hmmt2025archive,
hendrycks2021math,lightman2023lets,ge2026mathvault}. Methods differ only in the
sampled-token
coefficient or the shared TV-OPD scale. Appendix~\ref{app:experimental-details}
provides the full configuration and checkpoint-selection protocol.

\subsection{Main Results}

\begingroup
\setlength{\abovecaptionskip}{4pt}
\setlength{\belowcaptionskip}{4pt}
\begin{table}[!htbp]
  \centering
  \caption{Validation accuracy (mean@16, \%). For each trained method, the
  selected checkpoint has the highest six-benchmark mean for
  that run. Entries report mean $\pm$ sample standard deviation across runs;
  Student is the step-0 point estimate. Avg. is first computed within each
  run. Bold and underline mark the best and second-best trained methods within
  each model pair.}
  \label{tab:main-results}
  \scriptsize
  \setlength{\tabcolsep}{0.8pt}
  \renewcommand{\arraystretch}{1.08}
  \begin{tabular}{@{}lccccccc@{}}
  \toprule
  \multirow{2}{*}{\textbf{Method}}
  & \multicolumn{5}{c}{\textbf{Competition Math}}
  & \multicolumn{1}{c}{\textbf{General Math}}
  & \multirow{2}{*}{\textbf{Avg. $\uparrow$}} \\
  \cmidrule(lr){2-6}\cmidrule(lr){7-7}
  & \textbf{AIME24} & \textbf{AIME25} & \textbf{AIME26}
  & \textbf{AMC23} & \textbf{HMMT25} & \textbf{MATH500} & \\
  \midrule
  \rowcolor{black!6}
  \multicolumn{8}{@{}l}{\textbf{JustRL-1.5B $\rightarrow$ DS-Distill-Qwen-1.5B}} \\
  \textbf{Student} & 27.50 & 25.83 & 20.83 & 72.50 & 15.00 & 81.20 & 40.48 \\
  \midrule
  Raw OPD & \underline{52.50 $\pm$ 1.5} & 37.19 $\pm$ 3.7
  & \textbf{39.80 $\pm$ 0.9} & 87.35 $\pm$ 1.5
  & 20.63 $\pm$ 1.5 & 86.30 $\pm$ 0.4
  & \underline{53.96 $\pm$ 0.2} \\
  ClipOPD & \textbf{52.82 $\pm$ 3.4} & \textbf{40.32 $\pm$ 1.6}
  & \underline{39.59 $\pm$ 0.3} & 85.47 $\pm$ 2.9
  & 20.21 $\pm$ 0.3 & 85.08 $\pm$ 0.7 & 53.91 $\pm$ 0.5 \\
  PowerOPD & 50.21 $\pm$ 0.9 & 38.65 $\pm$ 1.3
  & 37.61 $\pm$ 3.7 & 85.08 $\pm$ 0.6
  & \underline{21.77 $\pm$ 0.7} & 84.58 $\pm$ 1.6
  & 52.98 $\pm$ 0.4 \\
  vOPD & 50.31 $\pm$ 1.0 & 35.52 $\pm$ 1.3
  & 38.65 $\pm$ 0.1 & \underline{88.44 $\pm$ 0.4}
  & 20.73 $\pm$ 2.8 & \underline{87.23 $\pm$ 0.2}
  & 53.48 $\pm$ 0.8 \\
  \rowcolor{blue!9}
  \textbf{TV-OPD / Ours} & 49.17 $\pm$ 1.2
  & \underline{38.86 $\pm$ 0.1} & 38.75 $\pm$ 2.1
  & \textbf{88.60 $\pm$ 0.2} & \textbf{24.69 $\pm$ 0.7}
  & \textbf{87.75 $\pm$ 0.1} & \textbf{54.63 $\pm$ 0.3} \\
  \midrule
  \rowcolor{black!6}
  \multicolumn{8}{@{}l}{\textbf{Qwen3-8B $\rightarrow$ Qwen3-8B-SFT}} \\
  \textbf{Student} & 60.83 & 51.67 & 57.50 & 89.38 & 32.50 & 91.60 & 63.91 \\
  \midrule
  Raw OPD & 67.50 $\pm$ 4.1 & 60.21 $\pm$ 4.1
  & 57.61 $\pm$ 1.0 & \textbf{91.80 $\pm$ 1.7}
  & 33.03 $\pm$ 1.6 & 91.90 $\pm$ 0.1 & 67.01 $\pm$ 0.1 \\
  ClipOPD & 67.30 $\pm$ 0.6 & 56.77 $\pm$ 6.6
  & 61.57 $\pm$ 1.9 & 90.63 $\pm$ 1.8
  & \textbf{36.25 $\pm$ 3.2} & \underline{92.33 $\pm$ 1.4}
  & 67.47 $\pm$ 0.2 \\
  PowerOPD & 66.88 $\pm$ 0.0 & 59.48 $\pm$ 0.1
  & 58.96 $\pm$ 1.5 & 90.94 $\pm$ 0.9
  & \underline{35.21 $\pm$ 4.1} & 91.70 $\pm$ 0.5
  & 67.19 $\pm$ 0.4 \\
  vOPD & \textbf{68.13 $\pm$ 4.1} & \underline{60.84 $\pm$ 2.9}
  & \underline{61.77 $\pm$ 1.3} & \underline{91.57 $\pm$ 0.9}
  & 33.75 $\pm$ 3.2 & 91.13 $\pm$ 0.9
  & \underline{67.86 $\pm$ 0.1} \\
  \rowcolor{blue!9}
  \textbf{TV-OPD / Ours} & \underline{67.82 $\pm$ 1.3}
  & \textbf{61.04 $\pm$ 0.0} & \textbf{64.48 $\pm$ 3.1}
  & 91.41 $\pm$ 0.7 & 33.96 $\pm$ 0.3
  & \textbf{92.38 $\pm$ 0.1} & \textbf{68.51 $\pm$ 0.3} \\
  \bottomrule
\end{tabular}

\end{table}
\endgroup
\FloatBarrier

Table~\ref{tab:main-results} compares the methods across five competition-math
benchmarks and MATH-500. TV-OPD achieves the highest six-benchmark mean on
both model pairs. Relative to the untrained students, it raises the mean from
$40.48\%$ to $54.63\%$ on the JustRL pair and from $63.91\%$ to $68.51\%$
on the Qwen pair, corresponding to gains of 14.15 and 4.60 points,
respectively.

On the JustRL pair, TV-OPD reaches $54.63\pm0.3\%$, exceeding the strongest
baseline, Raw OPD ($53.96\pm0.2\%$), by 0.67 mean points. It leads on AMC
2023 ($88.60\%$), HMMT 2025 ($24.69\%$), and MATH-500 ($87.75\%$), while
ClipOPD leads on AIME 2024 and AIME 2025 and Raw OPD leads on AIME 2026. The
aggregate gain therefore comes from cross-benchmark balance rather than
dominance on every task.

On the larger Qwen pair, TV-OPD obtains $68.51\pm0.3\%$, improving over the
strongest baseline, vOPD ($67.86\pm0.1\%$), by 0.65 mean points. It leads on
AIME 2025, AIME 2026, and MATH-500; vOPD leads on AIME 2024, Raw OPD on AMC
2023, and ClipOPD on HMMT 2025. Given the overlapping variability across several
benchmarks, the narrow aggregate margin should be
interpreted as competitive performance rather than a statistically resolved
improvement.

\subsection{TV-OPD Optimization Dynamics}
\label{sec:training-dynamics}

\begin{figure}[!htbp]
  \centering
  \begin{minipage}[t]{0.49\textwidth}
    \centering
    \includegraphics[width=\linewidth]{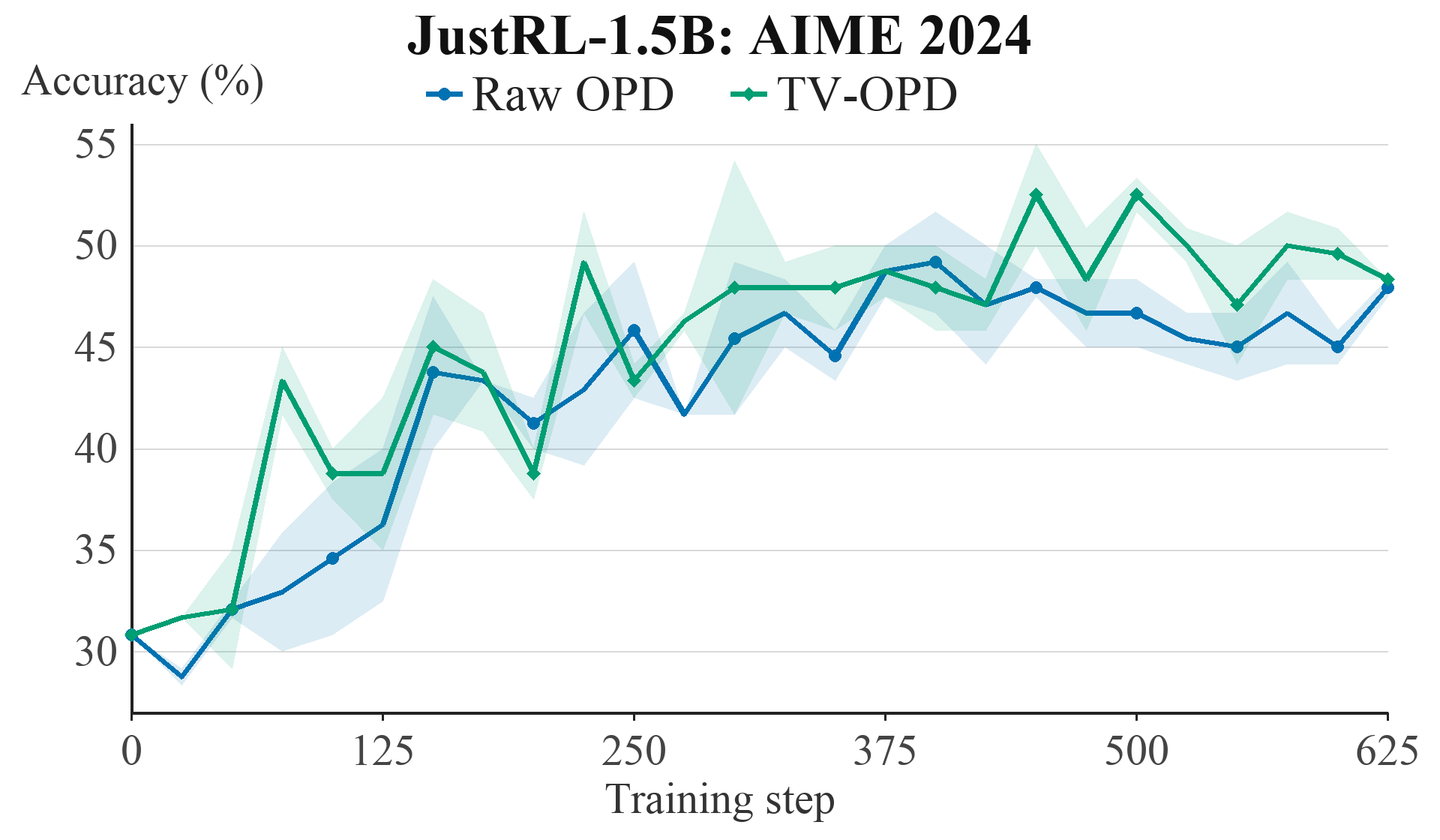}
    \textbf{(a) JustRL-1.5B, AIME 2024}
  \end{minipage}\hfill
  \begin{minipage}[t]{0.49\textwidth}
    \centering
    \includegraphics[width=\linewidth]{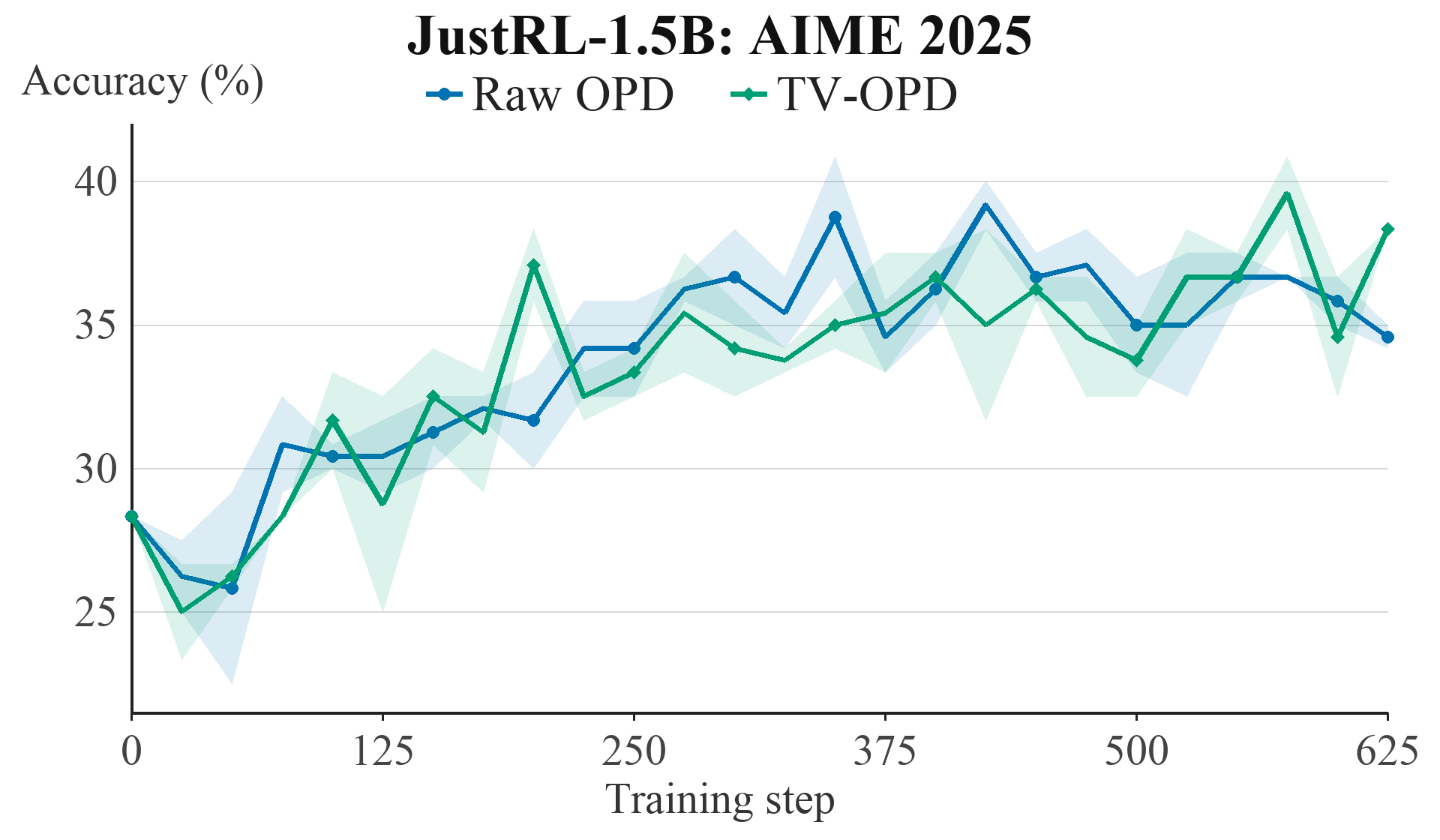}
    \textbf{(b) JustRL-1.5B, AIME 2025}
  \end{minipage}

  \vspace{2pt}
  \begin{minipage}[t]{0.49\textwidth}
    \centering
    \includegraphics[width=\linewidth]{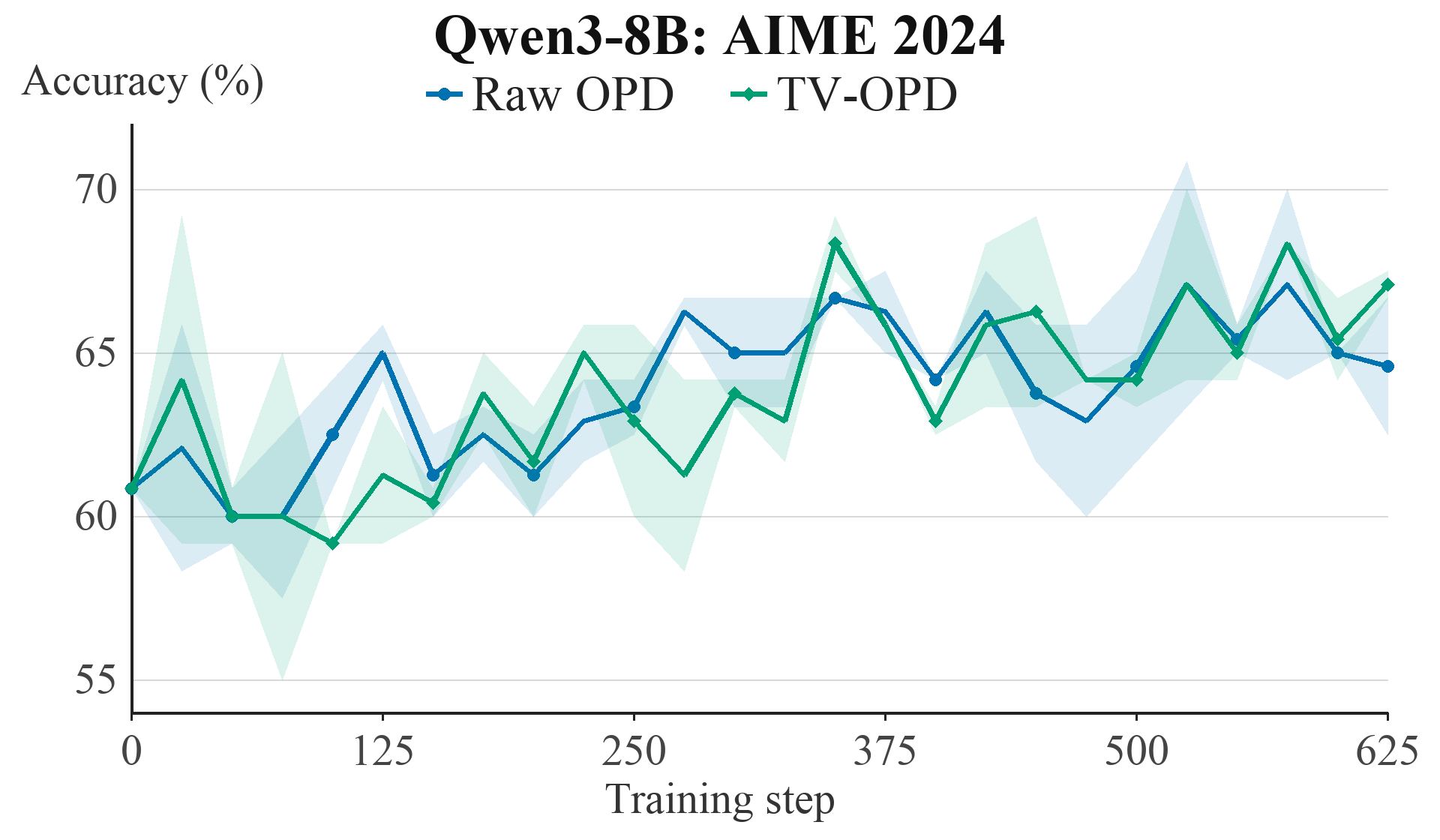}
    \textbf{(c) Qwen3-8B, AIME 2024}
  \end{minipage}\hfill
  \begin{minipage}[t]{0.49\textwidth}
    \centering
    \includegraphics[width=\linewidth]{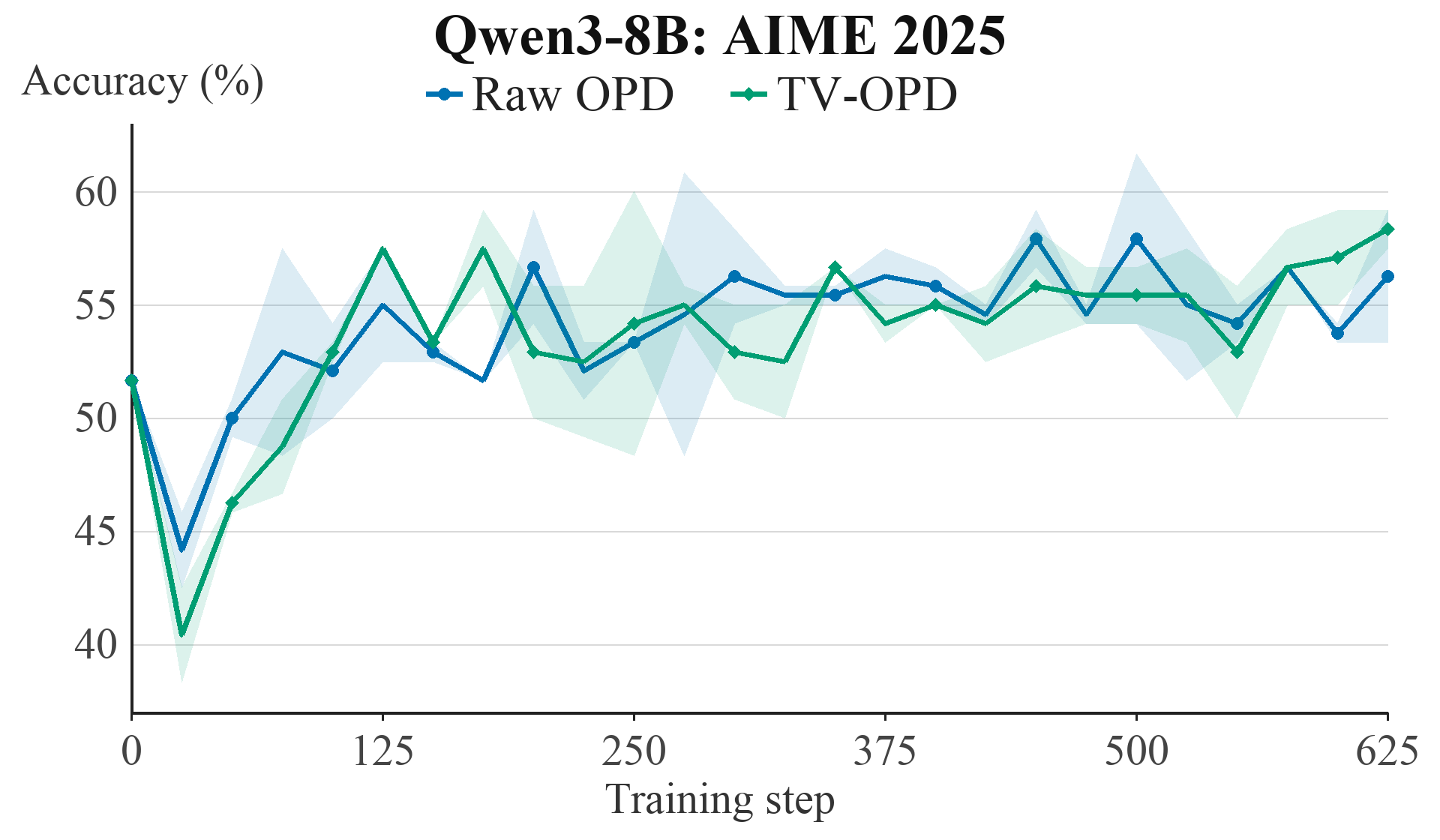}
    \textbf{(d) Qwen3-8B, AIME 2025}
  \end{minipage}

  \caption{Validation trajectories for Raw OPD and TV-OPD (mean $\pm$ one
  standard deviation across runs): JustRL-1.5B (a--b) and Qwen3-8B
  (c--d).}
  \label{fig:training-dynamics}
\end{figure}

Final checkpoint accuracy does not distinguish early learning from late-stage
retention. We therefore track AIME 2024, AIME 2025, and their average
\begin{equation}
  S_k=\frac{\mathrm{AIME24}_k+\mathrm{AIME25}_k}{2}.
  \label{eq:aggregate-score}
\end{equation}
The available exports contain validation accuracy but not the TV-OPD scale
$c_k$, conditional TV, or the realized update norm $\|\Delta\theta_k\|$. We
therefore analyze observed retention separately from these mechanism
diagnostics.
Figure~\ref{fig:training-dynamics} reports the corresponding paired-run
trajectories for the JustRL and Qwen model pairs.

\FloatBarrier

\begingroup
\setlength{\intextsep}{6pt plus 1pt minus 1pt}
\setlength{\abovecaptionskip}{4pt}
\setlength{\belowcaptionskip}{4pt}
\begin{table}[!ht]
  \centering
  \caption{Stage-wise JustRL validation accuracy (\%), averaged within each
  window and run and then across runs. Early, middle, and late denote
  steps 0--250, 275--450, and 500--625; bold marks the best method in each row.}
  \label{tab:training-stage-metrics}
  \small
  \setlength{\tabcolsep}{4pt}
  \renewcommand{\arraystretch}{0.92}
  \begin{tabular}{lccc}
  \toprule
  \textbf{Metric} & \textbf{TV-OPD ($\alpha=0.5$)}
    & \textbf{Sign-TV} & \textbf{Raw OPD} \\
  \midrule
  AIME 2024, early stage & 39.58 & \textbf{40.11} & 37.50 \\
  AIME 2024, middle stage & \textbf{48.28} & 46.72 & 46.41 \\
  AIME 2024, late stage & \textbf{49.58} & 47.92 & 46.11 \\
  \midrule
  AIME 2025, early stage & 30.45 & \textbf{31.70} & 30.49 \\
  AIME 2025, middle stage & 35.21 & 35.99 & \textbf{36.72} \\
  AIME 2025, late stage & \textbf{36.53} & 35.56 & 35.63 \\
  \bottomrule
\end{tabular}

\end{table}
\endgroup

Table~\ref{tab:training-stage-metrics} compares the three training stages.
Sign-TV performs best on both benchmarks in the early stage. In the middle
stage, TV-OPD leads on AIME 2024, while Raw leads on AIME 2025. In the late
stage, Raw and TV-OPD retain complete run coverage. TV-OPD is higher by 3.47 points on
AIME 2024 and 0.90 points on AIME 2025. The regulator does not improve every
stage; its clearest benefit is late-stage retention.

Using the late-stage retention metrics defined in
Appendix~\ref{app:training-curves}, TV-OPD improves LateMean from
$40.87\pm0.83$ to $43.06\pm0.10$ and reduces PeakDrop from $3.51\pm1.13$ to
$2.36\pm0.49$ among the methods with complete paired coverage.

The proposed mechanism can be tested by jointly tracking $c_k$, conditional
TV, and $\|\Delta\theta_k\|$. TV-OPD should reduce the global effective scale
without restoring token-wise $|\Delta_i|$.

\subsection{Sensitivity to Regulator Strength}

We isolate the global feedback strength on the Qwen pair by sweeping
$\alpha\in\{0, 0.25,0.5,1.0\}$ while leaving the local token weights and all
other training settings unchanged. Sign-TV provides the direction-only
reference without the TV-based global scale. We use the same six-benchmark
checkpoint-selection and mean@16 evaluation protocol as in
Table~\ref{tab:main-results}.
\begin{table}[!ht]
  \centering
  \caption{Expanded regulator-strength ablation on Qwen3-8B. Entries report
  mean@16 as mean $\pm$ sample standard deviation across runs at the selected
  checkpoint; Avg. is computed within each run over all six benchmarks. Bold
  marks the best result in each column.}
  \label{tab:alpha-ablation}
  \scriptsize
  \setlength{\tabcolsep}{0.9pt}
  \renewcommand{\arraystretch}{1.05}
  \begin{tabular}{@{}lccccccc@{}}
  \toprule
  \textbf{Method}
  & \textbf{AIME24} & \textbf{AIME25} & \textbf{AIME26}
  & \textbf{AMC23} & \textbf{HMMT25} & \textbf{MATH500}
  & \textbf{Avg.} \\
  \midrule
  Sign-TV
  & \textbf{69.2 $\pm$ 1.2} & 56.7 $\pm$ 2.4
  & 60.8 $\pm$ 3.5 & \textbf{92.5 $\pm$ 2.7}
  & 33.3 $\pm$ 3.5 & 92.3 $\pm$ 0.3 & 67.5 $\pm$ 0.6 \\
  TV-OPD ($\alpha=0.25$)
  & 65.5 $\pm$ 1.0 & 59.2 $\pm$ 3.8
  & 61.9 $\pm$ 0.9 & 92.2 $\pm$ 0.2
  & 35.0 $\pm$ 0.9 & 91.4 $\pm$ 0.1 & 67.5 $\pm$ 0.4 \\
  TV-OPD ($\alpha=0.5$)
  & 67.8 $\pm$ 1.3 & \textbf{61.0 $\pm$ 0.0}
  & \textbf{64.5 $\pm$ 3.1} & 91.4 $\pm$ 0.7
  & 34.0 $\pm$ 0.3 & 92.4 $\pm$ 0.1
  & \textbf{68.5 $\pm$ 0.3} \\
  TV-OPD ($\alpha=1.0$)
  & 66.8 $\pm$ 1.0 & 55.2 $\pm$ 2.4
  & 61.4 $\pm$ 0.7 & \textbf{92.5 $\pm$ 1.3}
  & \textbf{36.7 $\pm$ 1.8} & \textbf{92.7 $\pm$ 0.0}
  & 67.5 $\pm$ 0.2 \\
  \bottomrule
\end{tabular}

\end{table}

TV-OPD with $\alpha=0.5$ attains the highest six-benchmark average,
$68.5\pm0.3\%$, about 1.0 point above Sign-TV and both the
$\alpha=0.25$ and $\alpha=1.0$ variants. This gain is distributed across
AIME 2025, AIME 2026, and
MATH-500 rather than every task: Sign-TV leads AIME 2024, while
$\alpha=1.0$ leads HMMT 2025 and ties the best AMC 2023 score.

The expanded sweep therefore supports a moderate feedback strength rather
than monotonic improvement with stronger regulation. The lower average
standard deviation at $\alpha=1.0$ coincides with a lower mean, so stronger
attenuation does not improve the observed accuracy--stability trade-off.
These differences remain descriptive; we use $\alpha=0.5$ for the completed
comparisons and avoid
claiming a universal optimum across model scales.

\subsection{Case Study: Raw-Advantage Dispersion in a Training Batch}

\begin{figure}[!ht]
  \centering
  \includegraphics[width=0.98\textwidth]{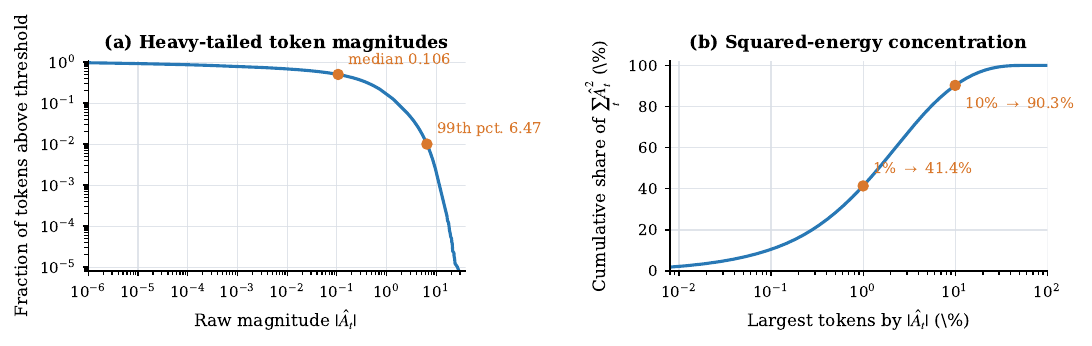}
  \caption{Raw-advantage statistics from one JustRL--DeepSeek-1.5B
  initialization batch (64 on-policy rollouts; 492,173 active response
  tokens). Left: empirical survival curve of token magnitudes. Right:
  cumulative squared energy after sorting tokens by magnitude.}
  \label{fig:raw-adv-training-snapshot}
\end{figure}

Figure~\ref{fig:raw-adv-training-snapshot} shows a pronounced mismatch between
typical and extreme raw coefficients. The pooled standard deviation is 1.34,
although the median $|\hat A_t|$ is only 0.106; the 99th percentile reaches
6.47 and the maximum reaches 34.58. Consequently, the largest 1\% of tokens
contribute 41.4\% of $\sum_t\hat A_t^2$, and the largest 10\% contribute
90.3\%. Every rollout contains at least one token with $|\hat A_t|>10$, so
the concentration is not attributable to a single anomalous sequence. This
single-batch case is descriptive rather than a population estimate, but it
confirms that highly uneven raw token weights arise in an actual OPD training
computation, supporting the magnitude-removal study in
Section~\ref{sec:observation}.



\section{Conclusion}
\label{sec:conclusion}

In this work, we show that sampled-token OPD can separate \emph{where to
update} from \emph{how strongly to train}. Sign-TV provides a principled
conditional TV objective, and TV-OPD adds a discrepancy-responsive global
scale without restoring token-wise magnitude. Across the tested settings,
this design remains competitive with raw OPD and improves late-stage retention
on the completed regulated comparison. These results provide a simple basis
for designing OPD objectives with explicit local direction and global control.

{\linespread{1}\fontsize{10}{11}\selectfont
\setlength{\bibsep}{8pt plus 1pt minus 1pt}
\bibliographystyle{iclr2027_conference}
\bibliography{references}
}

\clearpage
\appendix
\section{Theoretical Properties and Proofs}
\label{app:theory-proofs}

This appendix gives the formal statements supporting the sign-to-TV
derivation and the two properties summarized in Section~\ref{sec:method}:
bounded teacher-induced influence and behavioral preservation under TV
closeness. The estimator statement and its statistical analysis are in
Appendix~\ref{app:tv-estimation}.

\subsection{Reverse-KL sampled-token gradient}

For a fixed state $s$, abbreviate the student $q_\theta(a\mid s)$ by
$q_\theta(a)$ and the fixed teacher $p(a\mid s)$ by $p(a)$. Both have positive
probabilities on the finite vocabulary. Differentiation gives
\begin{align}
  \nabla_\theta D_{\mathrm{KL}}(q_\theta\|p)
  &=\sum_a\nabla_\theta q_\theta(a)
    \left(\log\frac{q_\theta(a)}{p(a)}+1\right)\\
  &=\mathbb E_{a\sim q_\theta}
  \left[\log\frac{q_\theta(a)}{p(a)}
  \nabla_\theta\log q_\theta(a)\right].
\end{align}
The second equality uses $\sum_a\nabla_\theta q_\theta(a)=0$.
Negating gives the expected update signal associated with the raw advantage
in Eq.~\eqref{eq:raw_advantage}. Averaging over the fixed state distribution,
the surrogate in Eq.~\eqref{eq:opd_objective}, with detached advantages,
reproduces this gradient at the on-policy point
$\theta=\bar\theta=\theta_{\mathrm{old}}$, where clipping is inactive.
Differentiating through state occupancy would introduce
additional terms and is outside this conditional objective.

\subsection{TV descent with bounded teacher influence}
\label{app:tv-gradient}

\begin{proposition}[TV descent with bounded teacher influence]
\label{prop:tv-gradient}
At a fixed state $s$, with a fixed teacher and positive token probabilities,
\[
  \mathbb E_{a\sim q_\theta}
  \left[\operatorname{sign}(\Delta(s,a))
  \nabla_\theta\log q_\theta(a\mid s)\right]
  =-2\nabla_\theta D_{\mathrm{TV}}(p,q_\theta)
\]
where TV is differentiable and $\theta=\bar\theta$. At equality coordinates,
$\operatorname{sign}(0)=0$ selects a generalized subgradient. For any shared
$c\ge0$, the token signal
$g_c=c\operatorname{sign}(\Delta)\nabla_\theta\log q_\theta$ satisfies
\begin{equation}
  |c\operatorname{sign}(\Delta)|\le c,\qquad
  \|g_c\|\le c\|\nabla_\theta\log q_\theta(a\mid s)\|.
  \label{eq:bounded-influence}
\end{equation}
\end{proposition}

\paragraph{TV gradient identity.}
At a differentiable point,
\begin{align}
  \nabla_\theta D_{\mathrm{TV}}(p,q_\theta)
  &=\frac12\sum_a\operatorname{sign}(q_\theta(a)-p(a))
    \nabla_\theta q_\theta(a)\\
  &=-\frac12\mathbb E_{a\sim q_\theta}
  \left[\operatorname{sign}(\Delta(a))
    \nabla_\theta\log q_\theta(a)\right].
\end{align}
Here $\Delta(a)=\log p(a)-\log q_\theta(a)$, and monotonicity of $\log$
identifies its sign with that of $p(a)-q_\theta(a)$. The teacher is fixed;
no gradient is taken through its probabilities or the coefficient.

At $q_\theta(a)=p(a)$, the subdifferential of $|q(a)-p(a)|$ with respect to
$q(a)$ is $[-1,1]$. Choosing zero on equality coordinates and the ordinary
sign elsewhere gives a valid subgradient in probability space. Composing
with the differentiable probability map yields a generalized subgradient in
parameter space. The resulting identity is a subgradient identity at such
points, not a claim of strict decrease for every finite step. Averaging over
fixed states gives the corresponding stopped-occupancy statement.

\subsection{Bounded teacher-induced influence and perturbations}
\label{app:bounded-influence}

Write $h(a)=\nabla_\theta\log q_\theta(a\mid s)$. For $\Delta(a)\ne0$,
\begin{equation}
  g^{\mathrm{TV}}(a)=\operatorname{sign}(\Delta(a))h(a)
  =\frac{g^{\mathrm{Raw}}(a)}{|\Delta(a)|}.
  \label{eq:direction-preservation}
\end{equation}
Thus every nonzero token direction is preserved, although their weighted sum
can change direction. Since $|\operatorname{sign}(\Delta)|\le1$, multiplying
by any $c\ge0$ proves Eq.~\eqref{eq:bounded-influence}. In comparison,
$|\log(p(a)/q_\theta(a))|$ has no uniform bound over positive distributions.

For two teachers $p_1,p_2$, hold the student, state, token, and global scalar
$c$ fixed. Their TV token signals obey
\begin{equation}
  \|g_c(p_1)-g_c(p_2)\|
  \le2c\|h(a)\|.
\end{equation}
The teacher may flip a token's direction but cannot induce arbitrarily large
relative weight at fixed $c$. This is a bounded-change statement, not
Lipschitz continuity in teacher probabilities: the sign can jump near
agreement. The bound concerns the multiplicative teacher coefficient; the
score norm remains a separate factor.

\subsection{Behavioral preservation under TV closeness}
\label{app:tv-behavior}

\begin{proposition}[Behavioral preservation under TV closeness]
\label{prop:tv-behavior}
For distributions $u,v$ at a common state and any statistic $f\in[m,M]$,
\begin{equation}
  |\mathbb E_u f-\mathbb E_v f|
  \le(M-m)D_{\mathrm{TV}}(u,v).
  \label{eq:tv-bounded-utility}
\end{equation}
In particular, the bound is $2\|f\|_\infty D_{\mathrm{TV}}(u,v)$ for bounded
$f$. If the current student's task advantage obeys
$|A^{q_\theta}(s,a)|\le A_{\max}(s)$ and the teacher has positive local margin
$\Delta_T(s)=\mathbb E_{a\sim p}A^{q_\theta}(s,a)>0$, then
\begin{equation}
  \mathbb E_{a\sim\tilde\pi} A^{q_\theta}(s,a)
  \ge\Delta_T(s)-2A_{\max}(s)D_{\mathrm{TV}}(\tilde\pi,p).
  \label{eq:local-improvement}
\end{equation}
Thus $D_{\mathrm{TV}}(\tilde\pi,p)<\Delta_T(s)/(2A_{\max}(s))$ preserves
positive expected advantage at this state.
\end{proposition}

\paragraph{Bounded statistics and the metric property.}
Let $P=\{a:u(a)\ge v(a)\}$. Because total signed mass is zero,
$\sum_{a\in P}(u(a)-v(a))=D_{\mathrm{TV}}(u,v)$.
For $f\in[m,M]$, centering by $m$ gives
\begin{align}
  \mathbb E_u f-\mathbb E_v f
  &=\sum_a(u(a)-v(a))(f(a)-m)\\
  &\le(M-m)\sum_{a\in P}(u(a)-v(a))
  =(M-m)D_{\mathrm{TV}}(u,v).
\end{align}
Swapping $u,v$ gives the absolute-value bound. Taking
$m=-\|f\|_\infty$ and $M=\|f\|_\infty$ yields its norm form.
The $\ell_1$ triangle inequality also gives, at the same state,
\begin{equation}
  D_{\mathrm{TV}}(q_\theta,\pi^\star)
  \le D_{\mathrm{TV}}(q_\theta,p)+D_{\mathrm{TV}}(p,\pi^\star).
  \label{eq:tv-triangle}
\end{equation}
These statements concern bounded next-action statistics at a common state.
A whole-response statistic instead requires a distance between whole-response
distributions; Appendix~\ref{app:sequence-tv} relates the two quantities.

\subsection{Conditional teacher-guided improvement}
\label{app:local-improvement}

Fix the current student's task advantage
$A^{q_\theta}(s,a)=Q^{q_\theta}(s,a)-V^{q_\theta}(s)$, under a specified
bounded-return task. This is an analysis quantity; TV-OPD does not estimate
it or use a task reward in its coefficient. Define
\begin{equation}
  \mathcal I_s(\pi)=\mathbb E_{a\sim\pi(\cdot\mid s)}A^{q_\theta}(s,a),
  \qquad \Delta_T(s)=\mathcal I_s(p)>0.
\end{equation}
With $A_{\max}(s)=\|A^{q_\theta}(s,\cdot)\|_\infty$, the norm form of
Eq.~\eqref{eq:tv-bounded-utility} implies
\begin{equation}
  \mathcal I_s(\tilde\pi)
  \ge\Delta_T(s)-2A_{\max}(s)D_{\mathrm{TV}}(\tilde\pi,p).
\end{equation}
A positive teacher margin implies $A_{\max}(s)>0$, so the strict neighborhood
condition in Proposition~\ref{prop:tv-behavior} ensures
$\mathcal I_s(\tilde\pi)>0$.

For comparison, the ideal mixture $\pi_\lambda=(1-\lambda)q_\theta+\lambda p$
with $0<\lambda\le1$ satisfies
\begin{equation}
  \mathcal I_s(\pi_\lambda)=\lambda\Delta_T(s)>0,\qquad
  D_{\mathrm{TV}}(\pi_\lambda,p)
  =(1-\lambda)D_{\mathrm{TV}}(q_\theta,p).
\end{equation}
This follows from $\mathcal I_s(q_\theta)=0$ and linearity. A neural-network
TV gradient step need not follow this mixture. Positive advantage at one
state is not a global return guarantee; extending it requires conditions
across visited states and control of occupancy shift. The teacher need not
be optimal, but a teacher-quality assumption is necessary for improvement.

\subsection{Support and tail behavior}

TV is finite and lies in $[0,1]$ even under support mismatch. Reverse KL is
infinite if $q(a)>0$ where $p(a)=0$. Under positive softmax probabilities,
arbitrarily small $p(a)$ still allows arbitrarily negative raw coefficients.
TV coefficients stay in $[-1,1]$. The sampled log-ratio identities assume
positive probabilities; changing the sampling distribution by top-$k$ or
nucleus truncation requires a separate support and importance-weighting
analysis. Bounded coefficients alone do not solve missing token coverage.



\section{On-Policy TV Estimation and Regulation}
\label{app:tv-estimation}

\subsection{Statewise estimator and exact variance}

\begin{proposition}[Bounded on-policy TV estimation]
\label{prop:tv-estimator}
Fix $s$ and normalized distributions with positive probabilities on the
shared vocabulary. For $a\sim q_{\bar\theta}(\cdot\mid s)$, define
\[
  \widehat d(s,a)
  =\left[1-\frac{p(a\mid s)}{q_{\bar\theta}(a\mid s)}\right]_+
  =[1-e^{\Delta(s,a)}]_+.
\]
Writing $d(s)=D_{\mathrm{TV}}(q_{\bar\theta}(\cdot\mid s),p(\cdot\mid s))$,
\begin{equation}
  \mathbb E[\widehat d\mid s]=d(s),\qquad
  0\le\widehat d\le1,\qquad
  \operatorname{Var}(\widehat d\mid s)\le d(s)(1-d(s))\le\tfrac14.
  \label{eq:tv-estimator-properties}
\end{equation}
\end{proposition}

\paragraph{Proof and equivalent estimators.}
Fix a state $s$ and abbreviate the rollout student $q_{\bar\theta}(\cdot\mid s)$
by $q$ and the teacher $p(\cdot\mid s)$ by $p$. Both are normalized and
positive on the finite shared vocabulary. All expectations below use the
same student that supplies the denominator of the likelihood ratio. Define
\begin{equation}
  X=\widehat d(s,a)=[1-p(a)/q(a)]_+,\quad a\sim q,\qquad
  d=d(s)=D_{\mathrm{TV}}(q,p).
\end{equation}
Normalization implies $\sum_a(q(a)-p(a))=0$, hence the total positive
and negative masses of $q-p$ agree. Therefore
\begin{align}
  \mathbb E[X\mid s]
  &=\sum_{a:q(a)>p(a)}q(a)\left(1-\frac{p(a)}{q(a)}\right)
    =\sum_{a:q(a)>p(a)}(q(a)-p(a))=d,\\
  \mathbb E[X^2\mid s]
  &=\sum_{a:q(a)>p(a)}\frac{(q(a)-p(a))^2}{q(a)},\\
  \operatorname{Var}(X\mid s)
  &=\sum_{a:q(a)>p(a)}\frac{(q(a)-p(a))^2}{q(a)}-d^2
    \le d(1-d)\le\frac14.
\end{align}
The variance bound follows from $0\le X\le1$, so $X^2\le X$. This proves
Proposition~\ref{prop:tv-estimator}. Positivity is a convenient sufficient
assumption for the log-ratio representation. The one-sided ratio identity
itself also holds when $q$ has zeros: sum only over $q(a)>0$, since any
coordinate with $q(a)>p(a)$ necessarily belongs to that support.

Under positive support, the alternatives $[p(a)/q(a)-1]_+$ and
$\frac12|1-p(a)/q(a)|$ also have mean $d$. Their expectations follow from the
negative mass of $q-p$ and the sum of its positive and negative masses,
respectively. They can be arbitrarily large when $p(a)/q(a)\gg1$; $X$ cannot.
If $q$ has zeros where $p$ has mass, those two alternative identities need
not hold, even though the one-sided identity for $X$ still does.
The equivalent evaluation $X=-\operatorname{expm1}(\min\{\Delta,0\})$
avoids exponentiating large positive log-ratios and reduces cancellation
near zero. This algebraic form is distinct from the direct exponential
expression used by the current implementation.

\subsection{Autoregressive sampling and the estimand}

Fix an optimizer step and condition on its pre-sampling history, so the
rollout policy $q=q_{\bar\theta}$ is fixed. For $b=1,\ldots,B$, draw independent
prompts $x_b\sim\rho$ and independent rollout randomness, set
$s_{b,t}=(x_b,a_{b,<t})$, and sample $a_{b,t}\sim q(\cdot\mid s_{b,t})$.
The teacher evaluates these same states. Training-history conditioning is
suppressed throughout the rollout calculations. Write
\begin{align}
  X_{b,t}&=\widehat d(s_{b,t},a_{b,t})=[1-e^{\Delta_{b,t}}]_+,\\
  d_{b,t}&=d(s_{b,t})
    =D_{\mathrm{TV}}(q(\cdot\mid s_{b,t}),p(\cdot\mid s_{b,t})),\qquad
  D_t^{\mathrm{pos}}=\mathbb E[d_{b,t}].
\end{align}
Here $D_t^{\mathrm{pos}}$ is the population TV at token position $t$;
$\widehat D_k$ later denotes a batch statistic at optimizer step $k$.
By the tower property, $\mathbb E X_{b,t}=\mathbb E d_{b,t}=D_t^{\mathrm{pos}}$.
For a deterministic horizon $H$,
\begin{equation}
  \widehat D_{B,H}=\frac1{BH}\sum_{b=1}^B\sum_{t=1}^H X_{b,t},\qquad
  \mathbb E\widehat D_{B,H}
  =\frac1H\sum_{t=1}^HD_t^{\mathrm{pos}}=D_H^{\mathrm{occ}}.
\end{equation}
Equivalently, writing $\nu_q^{\,t}$ for the rollout student's state
distribution at token position $t$, the target is
\begin{equation}
  D_H^{\mathrm{occ}}
  =\frac1H\sum_{t=1}^H\mathbb E_{s_t\sim\nu_q^{\,t}}
    D_{\mathrm{TV}}(q(\cdot\mid s_t),p(\cdot\mid s_t)).
  \label{eq:occupancy-tv}
\end{equation}
This is the fixed-horizon on-policy average TV described in the main text.
Token independence is unnecessary for this expectation.
Fixed-horizon results assume exactly $H$
defined positions; they do not treat a random EOS length as a fixed
normalizer. The i.i.d. rollout assumption also excludes dependent prompt
sampling; with fixed or dependent prompts, the target and variance must be
interpreted under that actual sampling design.

\subsection{Action-sampling noise and state-visitation noise}

Define the pre-action filtration
$\mathcal F_{b,t}=\sigma(x_b,a_{b,1},\ldots,a_{b,t-1})$, with the fixed
policy and teacher understood. Then $d_{b,t}$ is $\mathcal F_{b,t}$-measurable
and the residual $\varepsilon_{b,t}=X_{b,t}-d_{b,t}$ obeys
$\mathbb E[\varepsilon_{b,t}\mid\mathcal F_{b,t}]=0$. Thus
\begin{equation}
  X_{b,t}-D_t^{\mathrm{pos}}
  =\underbrace{\varepsilon_{b,t}}_{\text{action sampling}}
  +\underbrace{(d_{b,t}-D_t^{\mathrm{pos}})}_{\text{state visitation}}.
\end{equation}
For $t<u$, $\varepsilon_{b,t}$ is measurable with respect to
$\mathcal F_{b,u}$, so
$\mathbb E[\varepsilon_{b,t}\varepsilon_{b,u}]
=\mathbb E[\varepsilon_{b,t}\mathbb E(\varepsilon_{b,u}\mid\mathcal F_{b,u})]=0$.
Independence across rollouts gives orthogonality across $b$. Consequently,
\begin{equation}
  \operatorname{Var}\!\left(\frac1{BH}\sum_{b,t}\varepsilon_{b,t}\right)
  =\frac1{B^2H^2}\sum_{b,t}
    \mathbb E\operatorname{Var}(X_{b,t}\mid\mathcal F_{b,t})
  \le\frac1{4BH}.
\end{equation}
This is only the action-noise variance. An early action residual can correlate
with future $d_{b,u}$, so the total variance includes covariance between the
action and visitation components. No universal $O(1/(BH))$ bound follows
for $\widehat D_{B,H}$. Instead, each rollout mean
$Z_b=H^{-1}\sum_tX_{b,t}$ lies in $[0,1]$. Since the $Z_b$ are independent,
\begin{equation}
  \operatorname{Var}(\widehat D_{B,H})
  =\frac1{B^2}\sum_b\operatorname{Var}(Z_b)\le\frac1{4B}.
\end{equation}

A concentration bound follows directly from the conditional range of the
action residual. Given $\mathcal F_{b,t}$, it has mean zero and lies in
$[-d_{b,t},1-d_{b,t}]$, an interval of length one. The bounded-variable
exponential-moment inequality gives
$\mathbb E[e^{\lambda\varepsilon_{b,t}}\mid\mathcal F_{b,t}]
\le e^{\lambda^2/8}$.
Expose rollouts consecutively, adding each prompt before its actions; the
same conditional bound holds with previous rollouts included in the history.
Iterating it over $BH$ actions gives an exponential moment of at most
$e^{BH\lambda^2/8}$. Markov's inequality, optimized at
$\lambda=4u$ and applied to both signs, yields
\begin{equation}
  \Pr\!\left(\left|\frac1{BH}\sum_{b,t}(X_{b,t}-d_{b,t})\right|\ge u\right)
  \le2e^{-2BH u^2},\qquad u>0.
\end{equation}
This controls deviation from realized conditional TV, not from its population
mean $D_H^{\mathrm{occ}}$. For that target, applying the same bounded-variable
argument to the independent $Z_b$ gives only
$\Pr(|\widehat D_{B,H}-D_H^{\mathrm{occ}}|\ge u)\le2e^{-2Bu^2}$
without additional temporal assumptions.

\subsection{Variable length, EOS masks, and ratio estimation}
\label{app:tv-variable-length}

For rollouts capped at $H_{\max}$, let $\omega_{b,t}\in\{0,1\}$ indicate an
active token. Assume $\omega_{b,t}$ is $\mathcal F_{b,t}$-measurable, as when
the EOS token itself is included and only subsequent positions are excluded.
For formal convenience, extend each terminated trajectory with arbitrary
samples from $q$ up to the cap; their zero mask makes them irrelevant. Then
\begin{equation}
  \mathbb E[\omega_{b,t}X_{b,t}]
  =\mathbb E\!\left[\omega_{b,t}
    \mathbb E(X_{b,t}\mid\mathcal F_{b,t})\right]
  =\mathbb E[\omega_{b,t}d_{b,t}].
\end{equation}
Define $U_b=\sum_t\omega_{b,t}X_{b,t}$ and
$V_b=\sum_t\omega_{b,t}$. For i.i.d. rollouts and $\mathbb EV_b>0$, the target is
\begin{equation}
  D_{\mathrm{active}}
  =\frac{\mathbb EU_b}{\mathbb EV_b}
  =\frac{\sum_t\mathbb E[\omega_{b,t}d_{b,t}]}{\sum_t\mathbb E[\omega_{b,t}]},
  \qquad
  \widehat D_B^{\mathrm{active}}=\frac{\sum_bU_b}{\sum_bV_b}.
\end{equation}
The ratio is defined on batches with $\sum_bV_b>0$. Its numerator has the
correct expectation, but its random denominator prevents general
finite-sample unbiasedness. For example, suppose at the first position the
student assigns probability $1/2$ to EOS and $1/2$ to continuation, while the
teacher assigns $1/4$ and $3/4$. After continuation the two distributions
agree, and the cap is two. For one rollout, $(U,V)$ equals $(1/2,1)$ or
$(0,2)$, each with probability $1/2$. Hence
$\mathbb E[U/V]=1/4$, whereas $D_{\mathrm{active}}=(1/4)/(3/2)=1/6$.

Since $U_b,V_b$ are bounded, the law of large numbers gives
$\widehat D_B^{\mathrm{active}}\overset p\longrightarrow D_{\mathrm{active}}$
as $B\to\infty$ at a fixed policy. The event of an empty batch has probability
tending to zero when $\mathbb EV_b>0$; assigning any fixed bounded value on
that event does not change consistency. It can change finite-batch averages.
The exact expansion
\begin{equation}
  \sqrt B(\widehat D_B^{\mathrm{active}}-D_{\mathrm{active}})
  =\frac{B^{-1/2}\sum_b(U_b-D_{\mathrm{active}}V_b)}{B^{-1}\sum_bV_b}
\end{equation}
and the central limit theorem with Slutsky's theorem yield
\begin{equation}
  \sqrt B(\widehat D_B^{\mathrm{active}}-D_{\mathrm{active}})
  \Rightarrow\mathcal N\!\left(0,
  \frac{\operatorname{Var}(U_b-D_{\mathrm{active}}V_b)}{(\mathbb EV_b)^2}\right).
\end{equation}
A zero asymptotic variance is interpreted as a degenerate limit.
Thus conditional token estimates and fixed-horizon averages are exactly
unbiased; variable-length pooling is consistent under these assumptions.
Equal weighting of sequence means targets $\mathbb E[U_b/V_b]$ when
$V_b>0$ almost surely, a different quantity from pooled active-token TV.

\subsection{Relation to sequence-level TV}
\label{app:sequence-tv}

For deterministic $H$, define the conditional sequence distributions
$Q_x(\tau)=\prod_tq(a_t\mid s_t)$ and
$P_x(\tau)=\prod_tp(a_t\mid s_t)$. Let $Q,P$ be the joint prompt--response
laws with common prompt marginal $\rho$. The marginal cancels in their
likelihood ratio. The statewise one-sided argument, now on sequences, gives
\begin{equation}
  D_{\mathrm{TV}}(Q,P)
  =\mathbb E_{x\sim\rho,\,\tau\sim Q_x}
    \left[1-\exp\!\left(\sum_{t=1}^H\Delta_t\right)\right]_+.
\end{equation}
For a fixed prompt, the same identity holds for $Q_x,P_x$ without averaging
$x$. This bounded unbiased sequence-TV estimator differs from the average
of conditional token estimates. For $r_t=p(a_t\mid s_t)/q(a_t\mid s_t)$,
\begin{equation}
  \left[1-\prod_tr_t\right]_+\le\sum_t[1-r_t]_+.
\end{equation}
Indeed, replace $r_t$ by $\min\{r_t,1\}$ to decrease the product, then use
$1-\prod_t(1-u_t)\le\sum_tu_t$ for $u_t\in[0,1]$. Taking expectation under
$Q$, with the same prompt distribution as in $D_t^{\mathrm{pos}}$, gives
\begin{equation}
  D_{\mathrm{TV}}(Q,P)
  \le\min\!\left\{1,\sum_tD_t^{\mathrm{pos}}\right\}
  =\min\{1,H D_H^{\mathrm{occ}}\}.
\end{equation}
Persistent local discrepancies can cause sequence TV to approach one as
length grows. For example, two distinct i.i.d. Bernoulli token policies have
constant local TV, while an event separating their empirical token
frequencies has probabilities approaching one and zero under the two laws.
On-policy average TV measures local mismatch without summing it over length.

\subsection{EMA and scheduler predictability}
\label{app:tv-scheduler}

The per-step statistic $\widehat D_k$ is the pooled active-token ratio above,
computed using logged rollout probabilities, even if student parameters
subsequently change during optimization. At the start of step $k$, the
stored $c_k$ is frozen for all microbatches. TV numerators and token counts
accumulate across microbatches and are summed across data-parallel workers
before taking their ratio. This reproduces token pooling, not an unweighted
average of microbatch means.

The scheduler uses one EMA with retention $\beta=0.95$:
$\bar D_k=\beta\bar D_{k-1}+(1-\beta)\widehat D_k$ on accepted steps.
The first accepted step $k_0$ initializes
$\bar D_{k_0}=D_{\mathrm{ref}}=\widehat D_{k_0}$; the reference then stays
fixed. Initially $c_k=1$, and the first computed ratio is one as well.
The $\epsilon>0$ in Eq.~\eqref{eq:scheduled-tv-advantage} keeps the ratio
defined even if the empirical reference is zero. The new EMA sets only
$c_{k+1}$. A nonfinite gradient norm freezes the scheduler state.

For the statistical effect of smoothing, first consider an ideal sequence
without rejected steps, indexed from an initialized $\bar D_0$. Iteration gives
\begin{equation}
  \bar D_k=\beta^k\bar D_0
  +(1-\beta)\sum_{j=1}^k\beta^{k-j}\widehat D_j.
\end{equation}
Thus, with $\mu_j=\mathbb E\widehat D_j$, its expectation is the same weighted
sum of $\mathbb E\bar D_0$ and $\mu_j$; independence is not required for
this linear expectation. If the estimates are independent with constant mean
$D$ and variance $\sigma^2$, and independent of a finite-variance initial
value, summing the geometric variance series gives
\begin{equation}
  \mathbb E\bar D_k\longrightarrow D,\qquad
  \operatorname{Var}(\bar D_k)\longrightarrow
    \frac{1-\beta}{1+\beta}\sigma^2.
\end{equation}
The input mean is the mean of the batch estimator, which can itself carry
finite-batch ratio bias. These limits do not assert stationarity or
independence during adaptive training.

To describe skipped steps exactly after initialization, let $J_k\in\{0,1\}$
be the scheduler acceptance indicator and $\eta_k=(1-\beta)J_k$. Then
\begin{equation}
  \bar D_k=(1-\eta_k)\bar D_{k-1}+\eta_k\widehat D_k.
\end{equation}
If $J_k$ depends on current gradients, it can depend on the current data.
The weights are then random and correlated with the observations; the
fixed-weight expectation and stationary variance formulas cannot simply be
applied at the original optimizer-step indices.

Let $\mathcal H_{k-1}$ contain everything available before sampling step $k$,
including its rollout parameters and initialized reference. Since $c_k>0$
is measurable with respect to this history, for integrable gradient signals
\begin{equation}
  \mathbb E[c_k\widehat g_k^{\mathrm{TV}}\mid\mathcal H_{k-1}]
  =c_k\mathbb E[\widehat g_k^{\mathrm{TV}}\mid\mathcal H_{k-1}].
\end{equation}
When the base signal unbiasedly estimates the conditional-TV descent
direction under the chosen token weighting, this preserves its expected
direction. Fixed-horizon normalization provides such a setting;
random-denominator pooling retains its finite-sample qualification.
The identity concerns the sampled gradient signal before any
current-gradient-dependent rejection. A same-batch coefficient would instead
introduce a componentwise conditional covariance between $c_k$ and the
unscaled signal. Detaching it in autodifferentiation would not remove that
statistical dependence; historical scaling avoids it.

\subsection{Implementation scope: preprocessing and selection}
\label{app:tv-implementation}

The implementation evaluates $[1-\exp(\texttt{advantages}_i)]_+$ before
magnitude ablations, after optional chunk credit assignment and clamping.
The initial advantage is the teacher log-probability minus the rollout
student's log-probability. The effective mask is the response mask with
sentinel positions excluded. Its sum/count implementation divides by
$\max\{N_k,1\}$, where $N_k$ is the effective token count; an empty batch
therefore supplies zero, not a defined estimate of active-token TV. The
nonempty-batch assumption is needed when identifying a scheduler input with
that TV target. Including such zeros in the EMA changes its input mean.

Let $T(\Delta)$ denote optional preprocessing. Exact tokenwise agreement with
the TV estimator requires
$[1-e^{T(\Delta)}]_+=[1-e^\Delta]_+$ on sampled tokens. Equivalently,
$T(\Delta)=\Delta$ for $\Delta<0$ and $T(\Delta)\ge0$ for $\Delta\ge0$.
For example, capping positive log-ratios at a nonnegative threshold preserves
the statistic; changing the negative branch generally does not. These are
conditions for equality of the estimator itself, not a claim that all other
transformations must have biased expectation for every possible policy pair.

For token-dependent selection, consider a deterministic mask
$w(s,a)\in\{0,1\}$ at an otherwise active state. Its exact conditional
numerator and count are
\begin{equation}
  \mathbb E[w(s,a)X\mid s]=\sum_a w(s,a)[q(a\mid s)-p(a\mid s)]_+,
  \qquad
  \mathbb E[w(s,a)\mid s]=\sum_a w(s,a)q(a\mid s).
\end{equation}
The first expression is generally not
$d(s)\mathbb E[w(s,a)\mid s]$. Pooling such selected tokens targets the
ratio of expected selected positive mass to expected selected token count,
not necessarily active-state TV or TV of renormalized selected policies.
Thus predictable exclusions admit the occupancy proof above; current-token
sentinel exclusions require this selection-aware interpretation unless
additional properties establish equality. Finally, if generation uses a
sampling law different from the denominator distribution, the one-sided
proof must be rederived using that actual law or appropriate weighting.

\section{Full Experimental Setup}
\label{app:experimental-details}

\subsection{Baselines}
\label{app:baselines}

Let $\Delta_t=\log \pi_T(\hat y_t\mid s_t)-
\log \pi_{\theta_{\mathrm{old}}}(\hat y_t\mid s_t)$ denote the raw
sampled-token log-ratio. The following methods are compared in
Table~\ref{tab:main-results}.

\paragraph{Student and Raw OPD.}
\textbf{Student} is the step-0 initialization and receives no OPD update.
\textbf{Raw OPD} uses $A_t=\Delta_t$ in the clipped importance-sampling
objective of Eq.~\eqref{eq:opd_objective}; it is the untransformed
sample-based reverse-KL baseline~\citep{lu2025opd}.

\paragraph{ClipOPD and PowerOPD.}
\textbf{ClipOPD} replaces the raw coefficient with
$A_t=\operatorname{clip}(\Delta_t,c_{\min},c_{\max})$, limiting the
contribution of large log-ratio outliers~\citep{wang2026demystifying}.
\textbf{PowerOPD}
applies the Box--Cox family to token probabilities. For exponent $\gamma>0$,
its stop-gradient coefficient is
$A_t=\pi_T(\hat y_t\mid s_t)^\gamma-
\pi_{\theta_{\mathrm{old}}}(\hat y_t\mid s_t)^\gamma$. It is bounded in
$[-1,1]$, preserves the sign of $\Delta_t$, and has the raw log-ratio as its
rescaled $\gamma\rightarrow0$ limit
\citep{zhao2026poweropd}.

\paragraph{vOPD.}
\textbf{vOPD} treats sampled-token OPD as a policy-gradient estimator and
subtracts a detached, state-dependent control-variate baseline given by the
negative token-level reverse KL. The centering leaves the expected gradient
unchanged while reducing estimation variance~\citep{oh2026vopd}.

All baselines share the teacher and student checkpoints, prompt source and
shuffling protocol, training budget, optimizer settings, and evaluation code
within a model pair.
Method-specific transforms and thresholds follow their respective baseline
configurations. Raw OPD, Sign-TV, and TV-OPD are
also used as controlled comparisons in the training-dynamics analysis;
Sign-TV is the direction-only coefficient
$A_t=\operatorname{sign}(\Delta_t)$ before the proposed historical TV scale
is applied.

\subsection{Datasets}
\label{app:datasets}

\paragraph{Supervised-fine-tuning data.}
The Qwen3-8B student is initialized by supervised fine-tuning on 400K
examples sampled from OpenThoughts. OpenThoughts is a public reasoning-data
project spanning mathematics, code, science, and general reasoning; the
OpenThoughts3 recipe scales to 1.2M teacher-generated examples
\citep{guha2025openthoughts}. We use only the stated 400K subset to construct
the initial student. The released DeepSeek-R1-Distill-Qwen-1.5B checkpoint is
used directly for the 1.5B student~\citep{deepseekai2025r1}; we do not repeat
its upstream distillation stage.

\paragraph{OPD prompt data.}
For the Qwen pair, the loader uses the full DeepMath-103K training split
(103,022 rows in the public release) and consumes the \texttt{question} field
as the OPD prompt. The release also provides a verifiable final answer,
difficulty score, hierarchical topic, and three DeepSeek-R1 solutions per
problem. It emphasizes difficult problems, primarily levels 5--9, and was
semantically decontaminated against common mathematical benchmarks
\citep{he2025deepmath}. These answers, difficulty labels, and solutions are
not used as rewards by our OPD objective. For the JustRL pair, we use the full
DAPO-Math-17K prompt set, a curated collection of approximately 17K
mathematical problems with rule-verifiable answers released with the DAPO
system~\citep{yu2025dapo}. Again, training consumes the prompts while teacher
log probabilities provide the only supervision signal.

\paragraph{Evaluation data.}
The three AIME sets contain the two 15-problem forms from each of 2024, 2025,
and 2026 (30 problems per year). AMC 2023 is the 40-problem public subset used
by our evaluation harness. Both originate from the Mathematical Association
of America competition series~\citep{maa2026amc}. HMMT 2025 contains the 30
individual problems from the February 2025 Algebra and Number Theory,
Combinatorics, and Geometry tests; the official archive provides both
problems and solutions~\citep{hmmt2025archive}. Public machine-readable
snapshots and per-source provenance for these competition sets are catalogued
by math-vault~\citep{ge2026mathvault}. MATH-500 is the 500-problem held-out
subset introduced by \citet{lightman2023lets}, sampled from the original
12,500-problem MATH benchmark~\citep{hendrycks2021math}. The competition sets
probe recent short-answer contest mathematics, whereas MATH-500 covers seven
broader subjects with worked-solution source data. No evaluation problem or
gold answer is included in the OPD loss.

\paragraph{Training.}
The loader shuffles the full OPD prompt split and uses batches of 64. Prompts
are left-truncated at 1,024 tokens. One on-policy response of at most 16,384
tokens is sampled per prompt at temperature 1.0 and top-$p=1.0$. Training uses
AdamW~\citep{loshchilov2019adamw} with learning rate $10^{-6}$, weight decay
0.01, gradient clipping at
1.0, a constant base schedule with warmup. Runs use bfloat16, gradient
checkpointing, fully sharded data parallelism, and sequence parallelism on
eight GPUs.

\paragraph{Evaluation.}
We periodically evaluate AIME 2024, AIME 2025, AIME 2026, AMC 2023, HMMT
2025, and MATH-500 at temperature 0.6, top-$p=0.95$, top-$k=20$, and maximum
length 31,744. The main comparison and regulator-strength ablation sample 16
responses per problem and report mean@16. Training-dynamics analyses retain
their separately stated mean@4 protocol. The paired-run Raw, Sign-TV, and
TV-OPD trajectory analyses use AIME 2024 and AIME 2025.
The separate magnitude diagnostic is described in
Appendix~\ref{app:magnitude-ablations}.

\paragraph{Checkpoint selection.}
For Tables~\ref{tab:main-results} and~\ref{tab:alpha-ablation}, each completed
run independently selects the earliest checkpoint with the highest unweighted
six-benchmark mean within steps 1--1000. All six benchmark scores are read
from that same checkpoint, and the tables report the mean and sample standard
deviation across runs. The student row is the untrained step-0 evaluation. Training curves,
stage-wise summaries, and retention metrics retain all observations and their
separately stated horizons.

\paragraph{Method controls.}
Raw and TV differ only by replacing $\Delta_i$ with
$\operatorname{sign}(\Delta_i)$. TV-OPD retains exactly the TV token signs
and changes a single shared pre-optimizer scale $c_k$ using
the global-scale form in Eq.~\eqref{eq:scheduled-tv-advantage}; the completed
Qwen and JustRL comparisons both use $\alpha=0.5$.
No task reward or verifier is mixed into these
OPD coefficients.

The advantage-scaling scheduler configuration uses EMA retention $\beta=0.95$,
$c_{\min}=0.1$, and a reference frozen at the first valid optimizer step. The
coefficient is fixed at step entry and refreshed only for the following step.
It multiplies the policy-gradient loss uniformly across microbatches before
AdamW. All reported Qwen TV-OPD results use $\alpha=0.5$.
Appendix~\ref{app:tv-scheduler} describes the update order, and
Appendix~\ref{app:tv-implementation} states the assumptions needed after
masking and optional preprocessing.

\subsection{Hyperparameters}
\begin{table}[h]
  \centering
  \caption{Training and evaluation hyperparameters.}
  \label{tab:hyperparameters}
  \begin{tabular}{lcc}
    \toprule
    \textbf{Hyperparameter} & \textbf{Qwen3-8B pair} & \textbf{JustRL-1.5B pair} \\
    \midrule
    OPD prompt set & DeepMath-103K & DAPO-Math-17K \\
    Prompt subset & Full training split & Full training split \\
    TV-OPD $\alpha$ & 0.5 & 0.5 \\
    Training rollouts per prompt & 1 & 1 \\
    Training temperature / top-$p$ & 1.0 / 1.0 & 1.0 / 1.0 \\
    Maximum prompt / response length & 1,024 / 16,384 & 1,024 / 16,384 \\
    Prompt batch size & 64 & 64 \\
    Optimizer & AdamW & AdamW \\
    Learning rate & $1\times10^{-6}$ & $1\times10^{-6}$ \\
    Base LR schedule / warmup & Constant / linear & Constant / linear \\
    Weight decay / gradient clipping & 0.01 / 1.0 & 0.01 / 1.0 \\
    Precision & bfloat16 & bfloat16 \\
    Hardware & 8 GPUs & 8 GPUs \\
    Evaluation interval & Periodic & Periodic \\
    Evaluation samples per problem & 4 & 4 \\
    Evaluation temperature / top-$p$ & 0.6 / 0.95 & 0.6 / 0.95 \\
    $\epsilon$ in TV-Based scheduler & 0.00001 & 0.00001 \\
    \bottomrule
  \end{tabular}
\end{table}

\section{Full Magnitude Ablations}
\label{app:magnitude-ablations}

All interventions write the raw coefficient as $\Delta_i=z_im_i$, where
$z_i=\operatorname{sign}(\Delta_i)$ and $m_i=|\Delta_i|$, and preserve the
teacher-relative sign. Raw retains $m_i$; Sign sets every nonzero magnitude to
one. Sequence-Constant replaces every token magnitude in a sequence by the
sequence mean $S$, retaining the token signs but removing within-sequence
relative-magnitude allocation. Shuffle reassigns relative magnitudes across
tokens within each sequence while retaining the original token signs and
sequence-level scale, thereby corrupting token--magnitude correspondence.

\subsection{JustRL diagnostic}

The main diagnostic uses the JustRL-DeepSeek-1.5B teacher and
DeepSeek-R1-Distill-Qwen-1.5B student on AIME 2024 and 2025. Run lengths
differ, so we compute comparative summaries over the
jointly complete evaluation horizon and draw later reduced-coverage
observations as dashed segments in
Figures~\ref{fig:relative-magnitude-ablation}
and~\ref{fig:sequence-scale-ablation}.

\begin{table}[h]
  \centering
  \caption{Per-run best-over-training accuracy within the common evaluation
  horizon (mean $\pm$ sample standard deviation across runs).
  Sign is highest on both benchmarks.}
  \label{tab:magnitude-ablation-best}
  \small
  \begin{tabular}{lcccc}
  \toprule
  \textbf{Benchmark} & \textbf{Raw} & \textbf{Sign} & \textbf{Sequence-Constant} & \textbf{Shuffle} \\
  \midrule
  AIME 2024 & 47.50 $\pm$ 0.00 & \textbf{50.00 $\pm$ 1.18} & 47.50 $\pm$ 1.18 & 46.67 $\pm$ 2.36 \\
  AIME 2025 & 35.83 $\pm$ 1.18 & \textbf{37.50 $\pm$ 0.00} & 36.25 $\pm$ 5.30 & 35.00 $\pm$ 0.00 \\
  \bottomrule
\end{tabular}

\end{table}

For a trajectory-level summary, we first average AIME 2024 and AIME 2025 at
each checkpoint, average checkpoints within each run, and then average across
runs. This gives $36.29\pm0.15$ for Sign, $35.40\pm0.11$ for Raw,
$34.67\pm1.47$ for Sequence-Constant, and $34.35\pm0.13$ for Shuffle. These
values are diagnostic rather than a basis for significance claims; their role
is to motivate the magnitude-free objective analyzed in
Section~\ref{sec:tv-theory}.

\subsection{Additional allocation and sign-group controls}

Sign-Mass/Raw-Allocation provides an additional allocation control. For
$i\in G_g$, it sets the coefficient magnitude to
\begin{equation}
  \widetilde m_i
  = \frac{m_i}{|G_g|^{-1}\sum_{j\in G_g}m_j},
  \qquad g\in\{+,-\}.
  \label{eq:sign-mass-raw-allocation}
\end{equation}
The total magnitude of each sign group is therefore $|G_g|$, as under Sign,
while normalized raw magnitudes determine allocation within that group. The
run configuration records this control as
\texttt{opd\_adv\_mode=sign\_mass\_raw\_alloc}.

We compare this control with Raw on the JustRL pair using paired runs. The
runs have complete AIME 2024 and AIME 2025 evaluations through step 425, so
all summaries use the common 0--425 horizon. Training uses DAPO-Math-17K
\citep{yu2025dapo},
batches of 64 prompts, one rollout per prompt, AdamW with learning rate
$10^{-6}$, and the evaluation protocol in
Appendix~\ref{app:experimental-details}.

\begin{table}[h]
  \centering
  \caption{Per-run best-over-training accuracy within the common 0--425
  evaluation horizon (mean $\pm$ sample standard deviation across runs). Bold
  marks the higher mean in each row.}
  \label{tab:sign-mass-raw-allocation}
  \small
  \begin{tabular}{lcc}
  \toprule
  \textbf{Benchmark} & \textbf{Raw} & \textbf{Sign-Mass/Raw-Allocation} \\
  \midrule
  AIME 2024 & \textbf{50.83 $\pm$ 1.18} & 48.33 $\pm$ 1.18 \\
  AIME 2025 & \textbf{40.42 $\pm$ 1.77} & 37.08 $\pm$ 1.77 \\
  \bottomrule
\end{tabular}

\end{table}

For the trajectory-level summary used in the main diagnostic, the
two-benchmark mean is $37.00\pm0.28$ for Sign-Mass/Raw-Allocation and
$36.94\pm0.00$ for Raw. The control slightly improves the average trajectory
but not the per-benchmark best-over-training values in
Table~\ref{tab:sign-mass-raw-allocation}.

\subsection{Regulator sweep}

For TV-OPD, the $\alpha$ sweep changes only the global feedback sensitivity
in Eq.~\eqref{eq:scheduled-tv-advantage}. Table~\ref{tab:alpha-ablation}
reports the full Qwen six-benchmark sweep under the main checkpoint-selection
and mean@16 evaluation protocol.

\section{Additional Training-Curve Details}
\label{app:training-curves}

Figure~\ref{fig:training-dynamics} reports the available JustRL and Qwen AIME
trajectories with paired run coverage over the displayed 0--625-step horizon.
Stage-level JustRL benchmark averages are reported in
Table~\ref{tab:training-stage-metrics}.

Using the aggregate score $S_k$ in Eq.~\eqref{eq:aggregate-score}, let
$W=\{500,525,\ldots,625\}$ denote the matched late-training window. We define
\begin{align}
  \mathrm{LateMean}
  &=\frac{1}{|W|}\sum_{k\in W}S_k,\\
  \mathrm{PeakDrop}
  &=\max_k S_k-\frac{1}{|W|}\sum_{k\in W}S_k.
  \label{eq:retention-metrics}
\end{align}
LateMean measures sustained late-stage performance. PeakDrop measures the gap
between the best observed score over steps 0--625 and that late-stage mean.
Both statistics retain all observations. Their values are:
\begin{center}
  \small
  \begin{tabular}{lcc}
  \toprule
  \textbf{Method} & \textbf{LateMean} & \textbf{PeakDrop} \\
  \midrule
  Raw OPD & 40.87 $\pm$ 0.83 & 3.51 $\pm$ 1.13 \\
  TV-OPD & 43.06 $\pm$ 0.10 & 2.36 $\pm$ 0.49 \\
  \bottomrule
\end{tabular}

\end{center}
Values are means $\pm$ sample standard deviations across runs.
Each run contributes six evaluations to LateMean. PeakDrop uses the highest
aggregate score observed over steps 0--625.

\end{document}